\def\year{2019}\relax
\documentclass[letterpaper]{article} 
\usepackage{aaai19}  
\usepackage{times}  
\usepackage{helvet}  
\usepackage{courier}  
\usepackage{url}  
\usepackage{graphicx}  
\usepackage{todonotes}
\usepackage{booktabs}       
\usepackage{nicefrac}       
\usepackage{microtype}      

\usepackage{wrapfig}

\usepackage{comment}

\usepackage{algorithm}

\usepackage{amsmath}
\usepackage{amsfonts}
\usepackage{amssymb}

\usepackage{tikz}
\usetikzlibrary{fit}
\usetikzlibrary{backgrounds}
\usetikzlibrary{positioning}
\usetikzlibrary{shapes,arrows}
\usepackage{caption}
\usepackage[position=b]{subcaption}

\usepackage{textcomp}
\usepackage{mdframed}

\usepackage[noend]{algpseudocode}
\makeatletter
\def\BState{\State\hskip-\ALG@thistlm}
\makeatother

\title{Combined Reinforcement Learning via Abstract Representations}

\author{
Vincent Fran\c{c}ois-Lavet \\ McGill University, Mila \\ vincent.francois-lavet@mcgill.ca 
\\ \\
{\bf \Large Doina Precup} \\ McGill University, Mila, DeepMind \\ dprecup@cs.mcgill.ca
 \And 
Yoshua Bengio \\ Universit\'e de Montreal, Mila \\ yoshua.bengio@mila.quebec
\\ \\
{\bf \Large Joelle Pineau} \\ McGill University, Mila, Facebook AI Research \\ jpineau@cs.mcgill.ca
 }

\begin{document}

\maketitle

\begin{abstract}
In the quest for efficient and robust reinforcement learning methods, both model-free and model-based approaches offer advantages. In this paper we propose a new way of explicitly bridging both approaches via a shared low-dimensional learned encoding of the environment, meant to capture summarizing abstractions. We show that the modularity brought by this approach leads to good generalization while being computationally efficient, with planning happening in a smaller latent state space. In addition, this approach recovers a sufficient low-dimensional representation of the environment, which opens up new strategies for interpretable AI, exploration and transfer learning.
\end{abstract}

\section{Introduction}
In reinforcement learning (RL), there are two main approaches to learn how to perform sequential decision-making tasks from experience. The first approach is the model-based approach where the agent learns a model of the environment (the dynamics and the rewards) and then makes use of a planning algorithm to choose the action at each time step. The second approach, so-called model-free, builds directly a policy or an action-value function (from which an action choice is straightforward).
For some tasks, the structure of the policy (or action-value function) offers more regularity and thus a model-free approach would be more efficient, whereas in other tasks it may be easier to learn the dynamics directly due to some structure of the environment in which case a model-based approach would be preferable.  In practice, 
it is possible to develop a combined approach that incorporates both strategies.

\begin{figure}[ht!]
 \centering
\scalebox{0.6}{
\begin{tikzpicture}[->,thick]
\small
\tikzstyle{main}=[circle, minimum size = 9mm, thick, draw =black!80, node distance = 12mm]
\tikzstyle{rr}=[rounded rectangle, rounded rectangle west arc=5pt, rounded rectangle east arc=50pt, minimum size = 7mm, thick, draw =black!80, node distance = 12mm]

\foreach \name in {0,...,2}
    \node[main, fill = red!25] (s\name) at (\name*5,1.5) {$s_\name$};
\foreach \name in {0,...,1}
    \node[rr, fill = red!25] (env\name) at (\name*5+2.5,1.5) {environment};
\foreach \name in {0,...,1}
    \node[main, fill = red!25] (a\name) at (\name*5+1.5,0.5) {$a_\name$};
\foreach \name in {0,...,2}{
    \node[rr, fill = {rgb:black,0.;green,1;blue,1;white,10},text width=1.2cm, align=center] (e\name) at (\name*5+0,0.2) {encoder};%
    }
\foreach \name in {0,...,1}{
    \node[rr, fill = blue!25,text width=2.2cm, align=center, minimum height=6.75em] (mb\name) at (\name*5+2.5,-1.15) {};
    \node[above] at (mb\name.south) {model-based};
    }
\foreach \name in {0,...,1}
    \node[rr, fill = blue!25,text width=1.2cm, align=center] (tr\name) at (\name*5+2.5,-1.4) {transition\\model};
\foreach \name in {0,...,1}
    \node[rr, fill = blue!25,text width=0.9cm, align=center] (rm\name) at (\name*5+2.5,-0.6) {reward\\model};
\foreach \name in {0,...,2}
    \node[main, fill = {rgb:black,0.;green,1;blue,1;white,10},text width=1cm, align=center] (x\name) at (\name*5,-1.2) {abstract\\state};
\foreach \name in {0,...,1}
    \node[main, fill = red!25, align=center] (r\name) at (\name*5+3.5,0.5) {$r_\name$};
\foreach \name in {0,...,2}
    \node[rr, fill = green!40,text width=1cm, align=center] (mf\name) at (\name*5,-2.5) {model-free};
\foreach \name in {0,...,2}
    \node[main, fill = green!40] (V\name) at (\name*5,-3.5) {$Q$};

\node[] (h3) at (3*5-3.3,0) {\huge $\ldots$};

\foreach \h in {0,...,2}
       {
        \path (s\h) edge (e\h);
        \path (e\h) edge (x\h);
        \path (x\h) edge[green!50!black] (mf\h);
        \path (mf\h) edge[green!50!black] (V\h);
        \ifthenelse{\h = 2}{}{
        \path (s\h) edge[red] (env\h);
        \path (a\h) edge[red] (env\h);
        \path (env\h) edge[red] (r\h);
        \path (x\h) edge[blue] (mb\h);
        \path (a\h) edge[blue] (mb\h);
        \path (rm\h) edge[blue] (r\h);
       		}
       }
\foreach \current/\next in {0/1,1/2}
       {
        \path (tr\current) edge[blue] (x\next);
        \path (env\current) edge[red] (s\next);
       }
\end{tikzpicture}
}
\caption
[Illustration of the CRAR architecture.]
{Illustration of the integration of model-based and model-free RL in the CRAR architecture, with a low-dimensional abstract state over which transitions and rewards are modeled. The elements related to the actual environment dynamics are in red (the state $s_t$, the action $a_t$ and the reward $r_t$). The model-free elements are depicted in green (value function $Q(s,a)$) while the model-based elements (transition model and reward model) are in blue. The encoder and the abstract state are shared for both the model-based and model-free approaches and are depicted in light cyan. Note that the CRAR agent can learn from any off-policy data (red circles).} 
\label{fig:CRAR_architecture}
\end{figure}
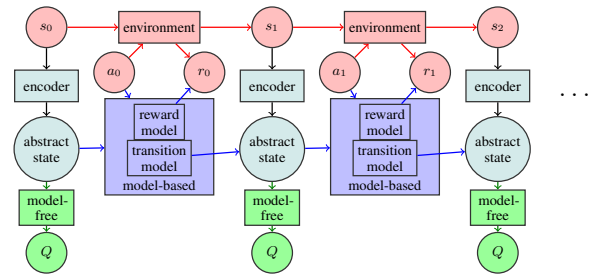

We present a novel deep RL architecture, which we call CRAR (Combined Reinforcement via Abstract Representations).
The CRAR agent combines model-based and model-free components, with the additional specificity that the proposed model forces both components to jointly infer a sufficient abstract representation of the environment. 
This is achieved by explicitly training both the model-based and the model-free components end-to-end, including the joint abstract representation.  To ensure the expressiveness of the abstract state, we also introduce an approximate entropy maximization penalty in the objective function, at the output of the encoder.
As compared to previous works that build implicitly an abstract representation through model-free objectives (see Section \ref{sec:related} for details), the CRAR agent creates a low-dimensional representation that captures meaningful dynamics, even in the absence of any reward (thus without the model-free part). In addition, our approach is modular thanks to the explicit learning of the model-based and model-free components.
The main elements of the CRAR architecture are illustrated in Figure~\ref{fig:CRAR_architecture}.

Learning everything through the abstract representation has the following advantages:
\begin{itemize}
\item it ensures that the features inferred in the abstract state provide good generalization, since they must be effective for both the model-free and the model-based predictions; 
\item it enables computationally efficient planning within the model-based module since planning is done over the abstract state space;
\item it facilitates interpretation of the decisions taken by the agent by expressing dynamics and rewards over the abstract state;
\item it allows developing new exploration strategies based on this low-dimensional representation of the environment.
\end{itemize}

In the experimental section, we show for two contrasting domains that the CRAR agent is able to build an interpretable low-dimensional representation of the task and that it can use it for efficient planning. We also show that the CRAR agent leads to effective multi-task generalization and that it can efficiently be used for transfer learning. 

\section{Formal setting}
We consider an agent interacting with its environment over discrete time steps. 
The environment is modeled as an MDP~\cite{bellman1957markovian}, defined by 
(i)~a state space $\mathcal S$ that is discrete or continuous;
(ii)~a discrete action space $\mathcal A=\{1, \ldots, N_{\mathcal A}\}$\label{ntn:N_A};
(iii)~the environment's transition function $T: \mathcal S \times \mathcal A \to \mathcal S $\label{ntn:trans_fct}, which we assume to be deterministic in this paper (although it can be extended to the stochastic case as discussed in Section~\ref{sec:discussion});
(iv)~the environment's reward function $R:~\mathcal S~\times~\mathcal A~\to~\mathcal R$\label{ntn:rwd_fct} where $\mathcal R$ is a continuous set of possible rewards in a range $R_{\text{max}} \in \mathbb{R}^+$\label{ntn:R_max} (e.g., $[0,R_{\text{max}}]$); and
(v)~a general discount factor $G:~\mathcal S~\times~\mathcal A~\times~\mathcal S~\rightarrow~[0, 1)$, similarly to \citeauthor{white2016unifying}~(\citeyear{white2016unifying})\footnote{The dependence of the discount factor on the transition is used for terminal states, where $G=0$. This is necessary so that the agent captures properly the implication of the end of an episode when planning (the cumulative future rewards equal to $0$ in a terminal state). 
Note that a biased discount factor $\Gamma: \mathcal S \times \mathcal A \times \mathcal S  \rightarrow [0, 1)$ is used during the training phase with $\Gamma \le G, \forall (s,a,s') \in (\mathcal S \times \mathcal A \times \mathcal S)$.}.
This setting encompasses the partially observable case if we consider that the state is a history of actions, rewards and observations.

The environment starts in a distribution of initial states $b(s_0)$.
At time step $t$, the agent chooses an action based on the state of the system $s_t \in \mathcal S$ according to a policy $\pi:\mathcal S \times \mathcal A \rightarrow [0,1]$.
After taking action $a_t \sim \pi(s_t,\cdot)$, the agent then observes a new state $s_{t+1} \in \mathcal S$ as well as a reward signal $r_t \in \mathcal R$ and a discount $\gamma_t \in \mathcal G$.
The objective is to optimize an expected return $V^\pi(s):\mathcal S \rightarrow \mathbb R$ such that
\begin{equation}
\small
V^\pi(s)=\mathbb E \left[ r_{t} + \sum \nolimits_{k=1}^{\infty} \Big(\prod_{i=0}^{k-1}\gamma_{t+i}\Big) r_{t+k} \mid s_t=s, \pi\right],
\label{def_V}
\end{equation}
where $r_{t} = \underset{a \sim \pi(s_t,\cdot)}{\mathbb E} R \big(s_{t},a \big)$, $\gamma_{t} = \underset{a \sim \pi(s_t,\cdot)}{\mathbb E} G \big(s_{t},a, s_{t+1} \big)$, and $s_{t+1}=T(s_{t},a_t)$.



\section{The CRAR agent}
We now describe in more detail the proposed CRAR approach illustrated in Figure~\ref{fig:CRAR_architecture}.

\subsection{CRAR components and notations}
We define an abstract state as $x \in \mathcal X$ where $\mathcal X = \mathbb R^{n_{\mathcal X}}$ and $n_{\mathcal X} \in \mathbb N$ is the dimension of the continuous abstract state space.
We define an encoder $e:\mathcal S \rightarrow \mathcal X$ as a function parametrized by $\theta_e$,
which maps the raw state $s$ to the abstract state $x$.
We also define the {\em internal} (or model) transition 
dynamics $\tau:\mathcal X \times \mathcal A \rightarrow \mathcal X$, parametrized by $\theta_\tau$: 
$
x'=x+\tau(x,a;\theta_\tau)
$.
In addition, we define the internal (or model) reward function $\rho:\mathcal X \times \mathcal A \rightarrow \mathcal R$, parametrized by $\theta_\rho$. 
For planning, we also need to fit the expected discount factor thanks to $g:\mathcal X \times \mathcal A \rightarrow [0,1)$, parametrized by $\theta_g$. 

In this paper, we investigate a model-free architecture with a Q-network $Q:\mathcal X \times \mathcal A \rightarrow \mathbb R$, parametrized by $\theta_Q$: 
$
Q(x, a; \theta_Q ),
$
which estimates
the expected value of discounted future returns.

\subsection{Learning the model}
Ideally a model-free learner uses an off-policy algorithm that can use past experience (with a replay memory) that is not necessarily obtained under the current policy. We use a variant of the DQN algorithm~\cite{mnih2015human}, called the double DQN algorithm~\cite{van2016deep}.
The current Q-value $Q(x, a; \theta_k )$ (for the abstract state $x$ relative to state $s$, when action $a$ is performed) is updated from a set of tuples $(s,a,r,\gamma,s')$ (with $r$ and $s'$ the observed reward and next-state), at every iteration, towards a target value:
\begin{equation}
\small
Y_k = r +\gamma Q\left(e(s';\theta_e^{-}),\operatorname*{argmax}_{a \in \mathcal A} Q(e(s';\theta_e),a;\theta_Q);\theta_Q^{-}\right),
\end{equation}
where, at any step $k$, $\theta_e^-$ and $\theta_Q^-$ are the parameters of earlier (buffered) encoder and Q-network, which together are called the target network.
The training is done by minimizing the loss
\begin{equation}
\mathcal L_{\text{mf}}(\theta_e, \theta_Q)=\left(Q(e(s;\theta_e), a; \theta_Q )-Y_k\right)^2.
\end{equation}
These losses are back-propagated into the weights of both the encoder and the Q-network. 
The model-free component of the CRAR agent could benefit in a straightforward way from using any other existing variant of DQN~\cite{hessel2017rainbow} or actor-critic architectures~\cite{mnih2016asynchronous}, where the latter would be able to deal with continuous action space or stochastic policies.


The model-based part is trained using data from the sequence of tuples $(s,a,r,\gamma,s')$. We have one loss for learning the reward, one for the discount factor\footnote{This could be extended to the current option in an option-critic architecture.},
and one for learning the transition\footnote{This loss is not applied when $\gamma=0$.}:
\begin{equation}
\mathcal L_\rho(\theta_e, \theta_\rho) = \mid r-\rho(e(s;\theta_e),a;\theta_\rho) \mid^2,
\end{equation}
\vspace{-0.3cm}
\begin{equation}
\mathcal L_g(\theta_e, \theta_g) = \mid \gamma-g(e(s;\theta_e),a;\theta_g) \mid^2,
\end{equation}
\vspace{-0.3cm}
\begin{equation}
\small
\mathcal L_\tau(\theta_e, \theta_\tau) = \mid (e(s;\theta_e)+\tau(e(s;\theta_e),a;\theta_\tau)-e(s';\theta_e)) \mid^2.
\end{equation}
These losses train the weights of both the encoder and the model-based components. These different components force the abstract state to represent the important low-dimensional features of the environment.
The model-based and the model-free approaches are complementary and both contribute to the abstract state representation. 

In practice, the problem that may appear is that a local minimum is found where too much information is lost in the representation $x=e(s;\theta_e)$. Keep in mind that if only the transition loss was considered, the optimal representation function would be a constant function (leading to 0 error in predicting the next abstract representation and a collapse of the representation). In practice the other loss terms prevent this but there is still a pressure to decrease the amount of information being represented (this will be clearly shown for the experiment described in Section \ref{sec:simplelaby} and in the ablation study in Appendix \ref{app:abla}). This loss of information mainly happens for states that are far (temporally) from any reward as the loss $L_\tau(\theta_e, \theta_\tau)$ then tends to keep the transitions trivial in the abstract state space. In order to prevent that contraction, a loss that encourages some form of entropy maximization in the state representation can be added. In our model, we use:
\begin{equation}
\mathcal L_{d1}(\theta_e) =  \text{exp}(- C_d \lVert e(s_1;\theta_e)-e(s_2;\theta_e)\rVert_2),
\end{equation}
where $s_1$ and $s_2$ are random states stored in the replay memory and $C_d$ is a constant.
As the successive states are less easily distinguished, we also introduce the same loss but with a particular sampling such that $s_1$ and $s_2$ are the successive states and we call it $\mathcal L'_{d1}$. Both losses are minimized.

The risk of obtaining very large values for the features of the state representation is avoided by the following loss that penalizes abstract states that are out of an $L_\infty$ ball of radius~$1$ (other choices are possible):
\begin{equation}
\mathcal L_{d2}(\theta_e) = \text{max}(  \lVert e(s_1;\theta_e)\lVert_\infty^2)-1,0).
\end{equation}
The loss $\mathcal L_{d}=\mathcal L_{d1}+\beta \mathcal L'_{d1}+\mathcal L_{d2}$ is called the representation loss and $\beta$ is a scalar hyper-parameter that defines the proportion of resampling of successive states for the loss $\mathcal L_{d1}$. 

At each iteration, a sum of the aforementioned losses are minimized using gradient descent\footnote{In practice, each term is minimized by mini-batch gradient descent (RMSprop in our case).}:
\begin{equation}
\begin{aligned}
\mathcal L=\alpha \big( \mathcal L_{\text{mf}}(\theta_e, \theta_Q) + \mathcal L_\rho(\theta_e, \theta_\rho) + \mathcal L_g(\theta_e, \theta_g)+ \\
\mathcal L_\tau(\theta_e, \theta_\tau) + \mathcal L_{d}(\theta_e) \big),
\end{aligned}
\end{equation}
where $\alpha$ is the learning rate. Details for the architecture and hyper-parameters used in the experiments are given in the appendix \ref{app:NN}\footnote{The source code for all experiments is available at \mbox{\url{https://github.com/VinF/deer/}}}. In the experiments, we will show the effect of the different terms, in particular how it is possible to learn an abstract state representation only from $L_\tau(\theta_e, \theta_\tau)$ and $\mathcal L_{d}$ in a case where there is no reward, we will discuss the importance of the representation loss $\mathcal L_{d}$ as well as the effect of $\alpha$ and $\beta$. 

\subsection{Interpretable AI}
In several domains it may be useful to recover an interpretable solution, which has sufficient structure to be meaningful to a human. Interpretability in this context could mean that some (few) features of the state representation are distinctly affected by some actions. To achieve this, we add the following optional loss (which will be used in some of the experiments) to make the predicted abstract state change aligned with the chosen embedding vector $v(a)$:
\begin{equation}
\mathcal L_{interpr}(\theta_e, \theta_\tau) = -\text{cos}\Big(\tau(e(s;\theta_e),a;\theta_\tau)_{0:n} , v(a) \Big),
\end{equation}
where $\text{cos}$ stands for the cosine similarity\footnote{Given two vectors $a$ and $b$, the cosine similarity is computed by
$\frac{a \cdot b}{(\|a\| \|b\|)+\epsilon}$ where $\epsilon$ is a small real number used to avoid division by $0$ when $\|a\|=0$ or $\|b\|=0$.
} and where the $n$-dimensional vector $v(a)$ ($n \in \mathbb N \le n_{\mathcal X}$) provides the direction that is softly encouraged for the $n$ first features of the transition in the abstract domain (when taking action $a$). The learning rate associated with that loss is denoted $\alpha_{interpr}$. The CRAR framework is sufficiently modular to incorporate other notions of interpretability; one could for instance think about maximizing the mutual information between the action $a$ and the direction of the transitions $\tau(e(s;\theta_e),a;\theta_\tau)$, with techniques such as MINE~\cite{belghazi2018mine}.

\subsection{Planning}
\label{sec:planning}
The agent uses both the model-based and the model-free approaches to estimate the optimal action at each time step.
The planning is divided into an expansion step and a backup step, similarly to \citeauthor{oh2017value}~(\citeyear{oh2017value}).  One starts from the estimated abstract state $\hat x_t$ and consider a number $b_d \le N_{\mathcal A}$ of best actions based on $Q (\hat x_t, a; \theta_Q)$ ($b_d$ is a hyper-parameter that simply depends on the planning depth $d$ in our setting). By simulating these $b_d$ actions with the model-based components, the agent reaches $b_d$ new different $\hat x_{t+1}$. For each of these $\hat x_{t+1}$, the expansion continues for a number $b_{d-1}$ of best actions, based on $Q (\hat x_{t+1}, a; \theta_Q)$. This expansion continues overall for a depth of $d$ expansion steps. During the backup step, the agent then compares the simulated trajectories to select the next action. 

We now formalize this process.
The dynamics for some sequence of actions is estimated recursively as follows for any $t'$:
\begin{equation}
    \hat x_{t'}=\left\{
                \begin{array}{ll}
                  e(s_t;\theta_e), \text{ if } t' = t\\
                  \hat x_{t'-1}+\tau(\hat x_{t'-1},a_{t'-1};\theta_\tau), \text{ if } t' > t
                \end{array}
              \right.
\end{equation}
We define recursively the depth-$d$ estimated expected return as
\begin{equation}
    \hat Q^d(\hat x_t, a)=\left\{
                \begin{array}{ll}
                  \rho(\hat x_t,a)+ g(\hat x_t,a) \ \underset{a' \in \mathcal A^*}{\operatorname{max}} \ \hat Q^{d-1}(\hat x_{t+1},a'), \\ \hspace{4cm} \text{ if } d > 0\\
                  Q(\hat x_t, a; \theta_k ), \text{ if } d = 0
                \end{array}
              \right.
\label{eq:Qd}
\end{equation}
where $\mathcal A^*$ is the set of $b_d$ best actions based on $Q (\hat x_t, a; \theta_Q)$ ($\mathcal A^* \subseteq \mathcal A$).
To obtain the action selected at time $t$, we use a hyper-parameter $D \in \mathbb N$ which quantify the depth of planning. We then use a simple sum of the Q-values obtained with planning up to a depth $D$: 
\begin{equation}
Q_{plan}^D(\hat x_t, a)=\sum_{d=0}^{D}  \hat Q^d(\hat x_t, a).
\end{equation}
The optimal action is given by
$\underset{a \in \mathcal A}{\operatorname{argmax}} \ Q_{plan}^D(\hat x_t, a)$.
Note that, using only $b_d$-best options at each expansion step is important for computational reasons.  Indeed, planning has a computational complexity that grows with the number of potential trajectories tested. In addition, it is also important to avoid overfitting to the model-based approach. Indeed, with a long planning horizon, the errors on the abstract states will usually grow due to the model approximation. When the internal model is accurate, a longer planning horizon and less pruning is beneficial, while if the model is inaccurate, one should rely less on planning.

\section{Experiments}
\label{sec:experiments}

\subsection{Labyrinth task}
\label{sec:simplelaby}
\begin{wrapfigure}{R}{3.8cm}
\vspace{-25pt}
  \centering
  \includegraphics[width=.99\linewidth]{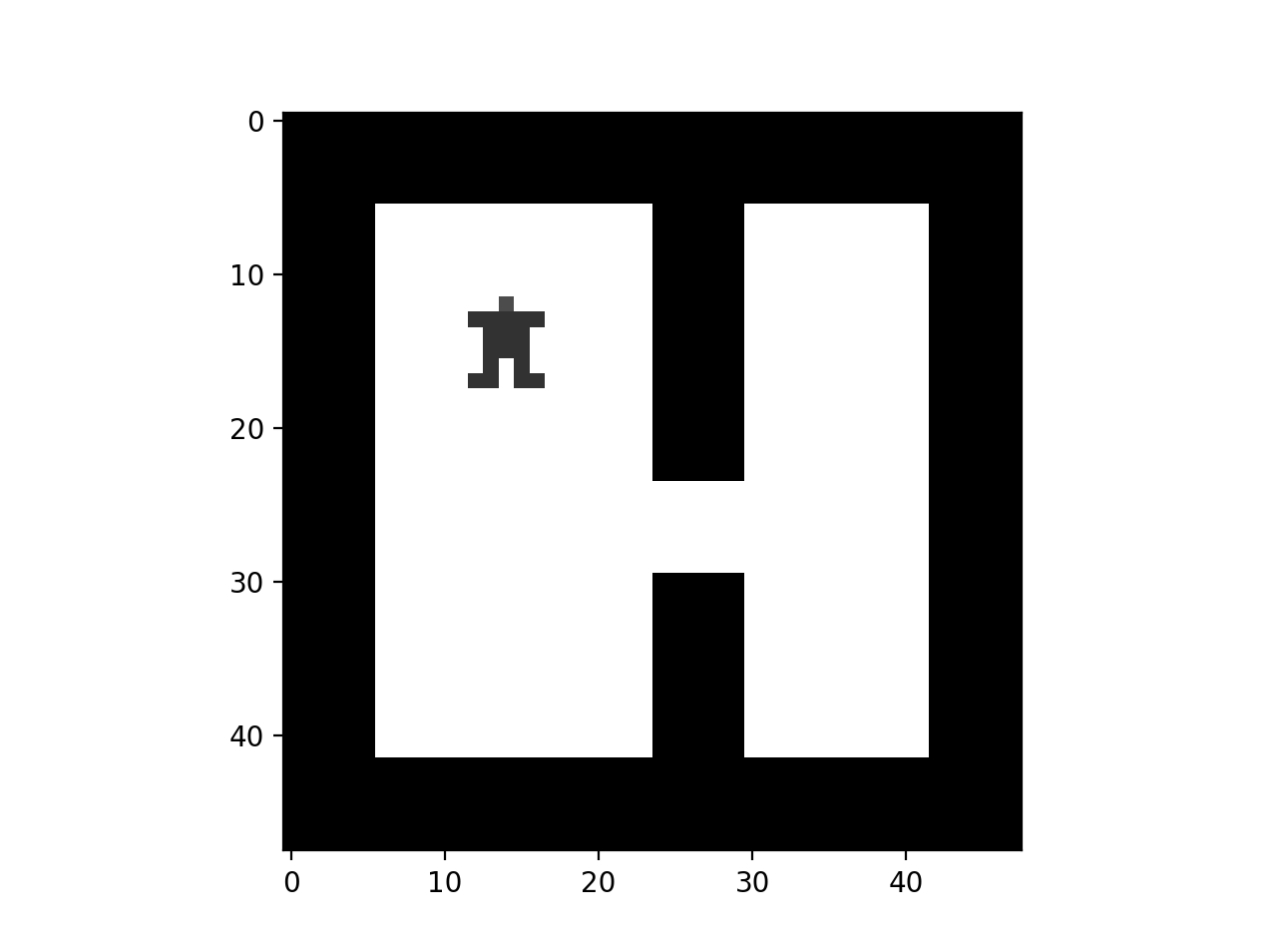}
  \caption{Representation of one state for a labyrinth task (without any reward).}
  \label{fig:lab_repr}
\vspace{-5pt}
\end{wrapfigure}

First, we consider a labyrinth MDP with four actions illustrated in Figure~\ref{fig:lab_repr}. The agent moves in the four cardinal directions (by 6 pixels) thanks to the four possible actions, except when the agent reaches a wall (block of $6 \times 6$ black pixels). This simple labyrinth MDP has no reward --- $r=0$, $\forall (s,a) \in (\mathcal S, \mathcal A)$ --- and no terminal state --- $\gamma=1, \forall (s,a) \in (\mathcal S, \mathcal A)$. As a consequence, the reward loss $\mathcal L_\rho$, the discount loss $\mathcal L_{g}$ and the model-free loss $\mathcal L_{\text{mf}}$ are trivially learned and can be removed without any noticeable change. 

\begin{figure}[ht!]
\centering
  \includegraphics[width=.55\linewidth]{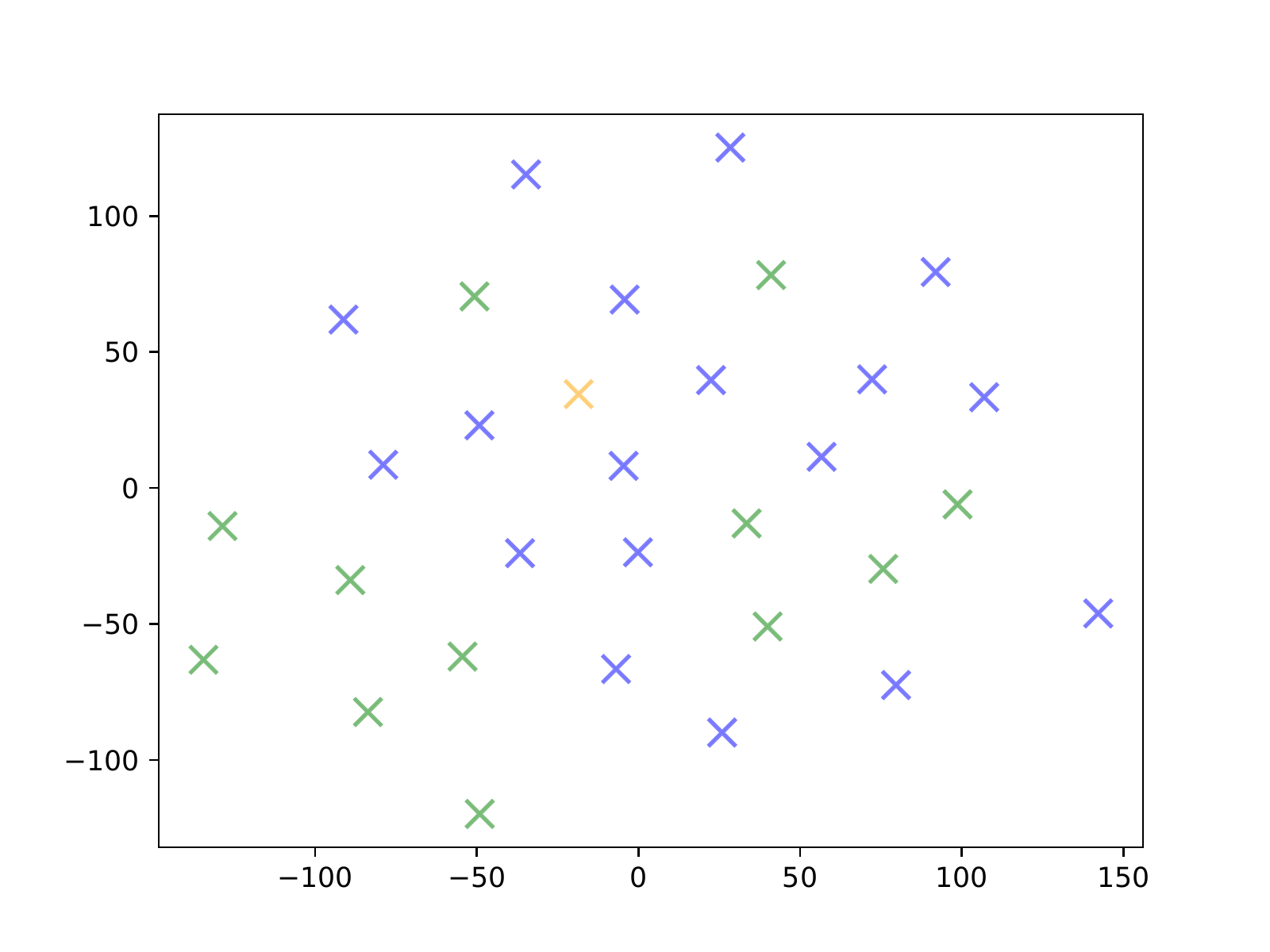}
  \caption{Two-dimensional representation of the simple labyrinth environment using t-SNE (blue represents states where the agent is on the left part, green on the right part and orange in the junction). This plot is obtained by running the t-SNE algorithm from a dataset containing all possible states of the labyrinth task. The perplexity used is 20.}
  \label{fig:tsne}
\end{figure}

\begin{figure}[ht!]
    \centering
    \begin{subfigure}[t]{0.21\textwidth}
        \centering
        \includegraphics[height=1.2in]{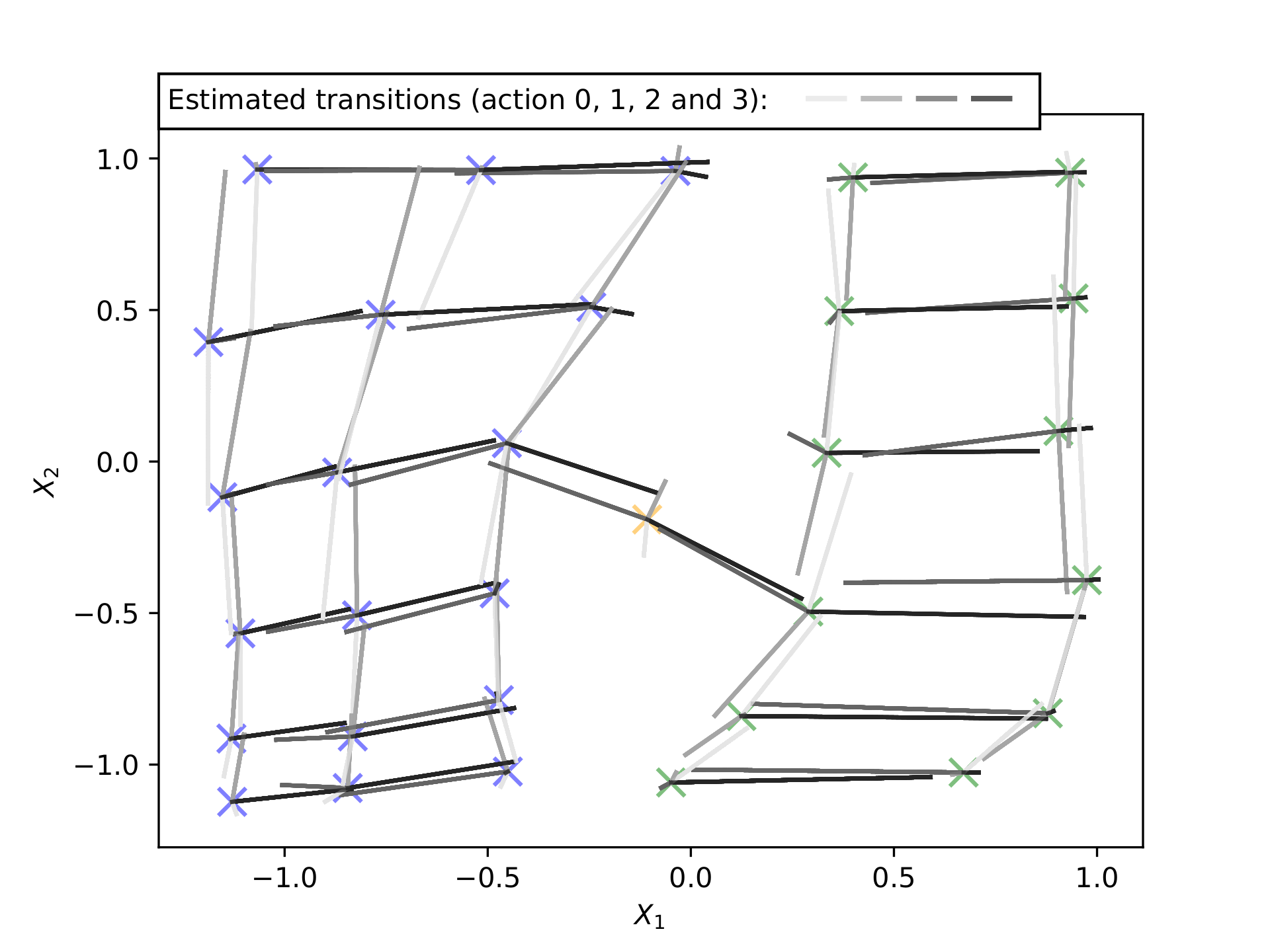}
        \caption{Without using the interpretability loss $\mathcal L_{interpr}$.}
        \label{fig:lab_dis}
    \end{subfigure}%
    \ \ \
    \begin{subfigure}[t]{0.21\textwidth}
        \centering
        \includegraphics[height=1.2in]{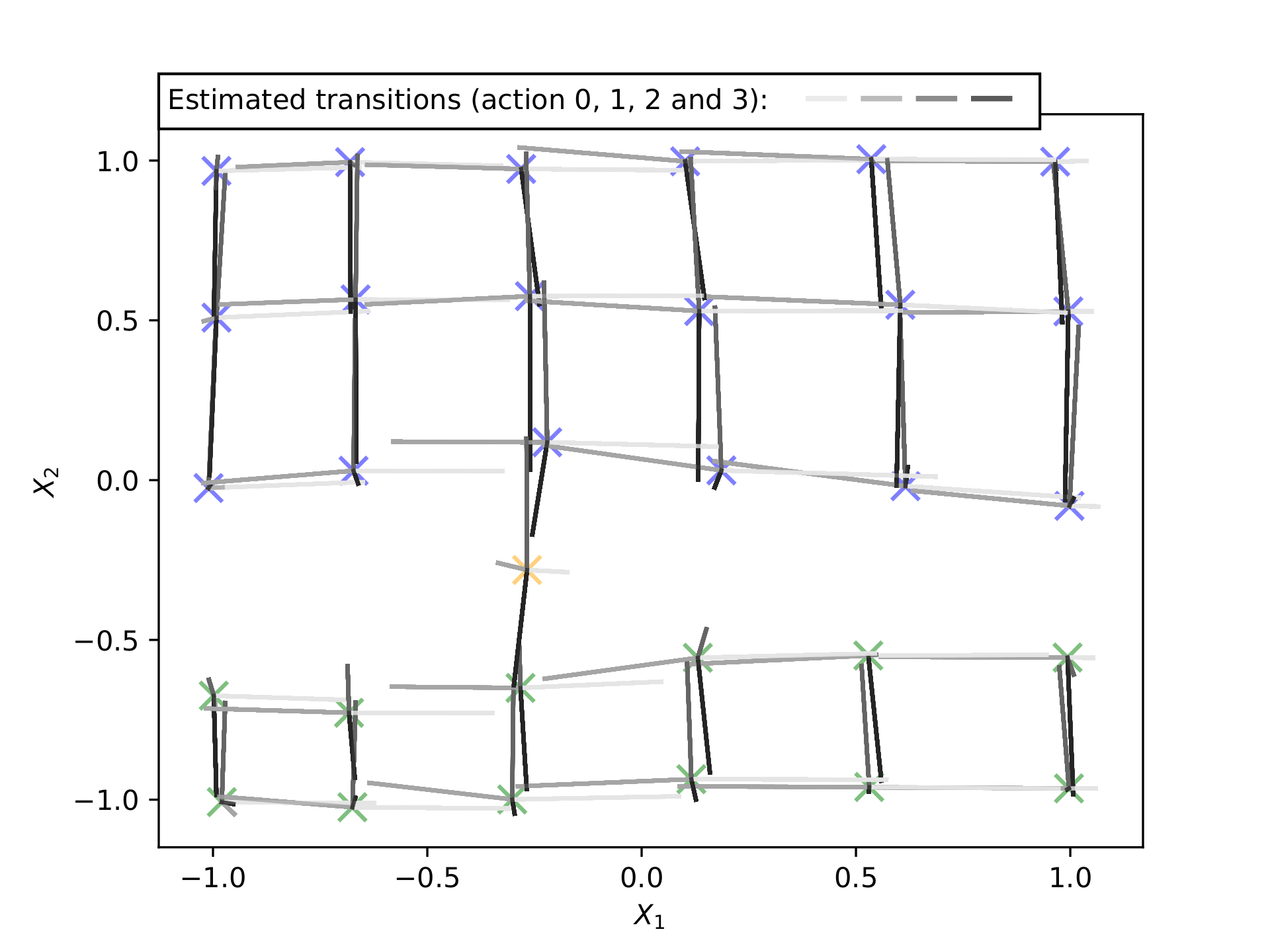}
        \caption{With enforcing $\mathcal L_{interpr}$ and $v(a_0)=[1,0]$, the action~$0$ is forced to correspond to an increasing feature $X_1$.}
        \label{fig:lab_forced}
    \end{subfigure}
    \caption{The CRAR agent is able to reconstruct a sensible representation of its environment in 2 dimensions.}
\end{figure}

As can be seen in Fig \ref{fig:tsne}, using techniques such as t-SNE~\cite{maaten2008visualizing} are inefficient to represent a meaningful low-dimensional representation of this task\footnote{The implementation used in Figure \ref{fig:tsne} can be found at the address \url{https://lvdmaaten.github.io/tsne/}}. This is because methods such as t-SNE or auto-encoders do not make use of the dynamics and only provide a representation based on the similarity between the visual inputs.
As opposed to this type of methods, we show in Figure~\ref{fig:lab_dis} that the CRAR agent is able to build a disentangled 2D abstract representation of the states. The dataset used is made up of 5000 transitions obtained with a purely random policy. Details, hyper-parameters along with an ablation study are provided in Appendix~\ref{app:laby}. This ablation study shows the importance of the representation loss $\mathcal L_{d}$ and it also shows that replacing the representation loss $\mathcal L_{d}$ by a reconstruction loss (via an auto-encoder) is not suitable to ensure a sufficient diversity in the low-dimensional abstract space.


In addition, when adding $\mathcal L_{interpr}$, it is shown in Figure~\ref{fig:lab_forced} how forcing some features can be used for interpretable AI.

\subsection{Catcher}
\begin{wrapfigure}{R}{3.5cm}
\vspace{-25pt}
  \centering
  \includegraphics[width=.99\linewidth]{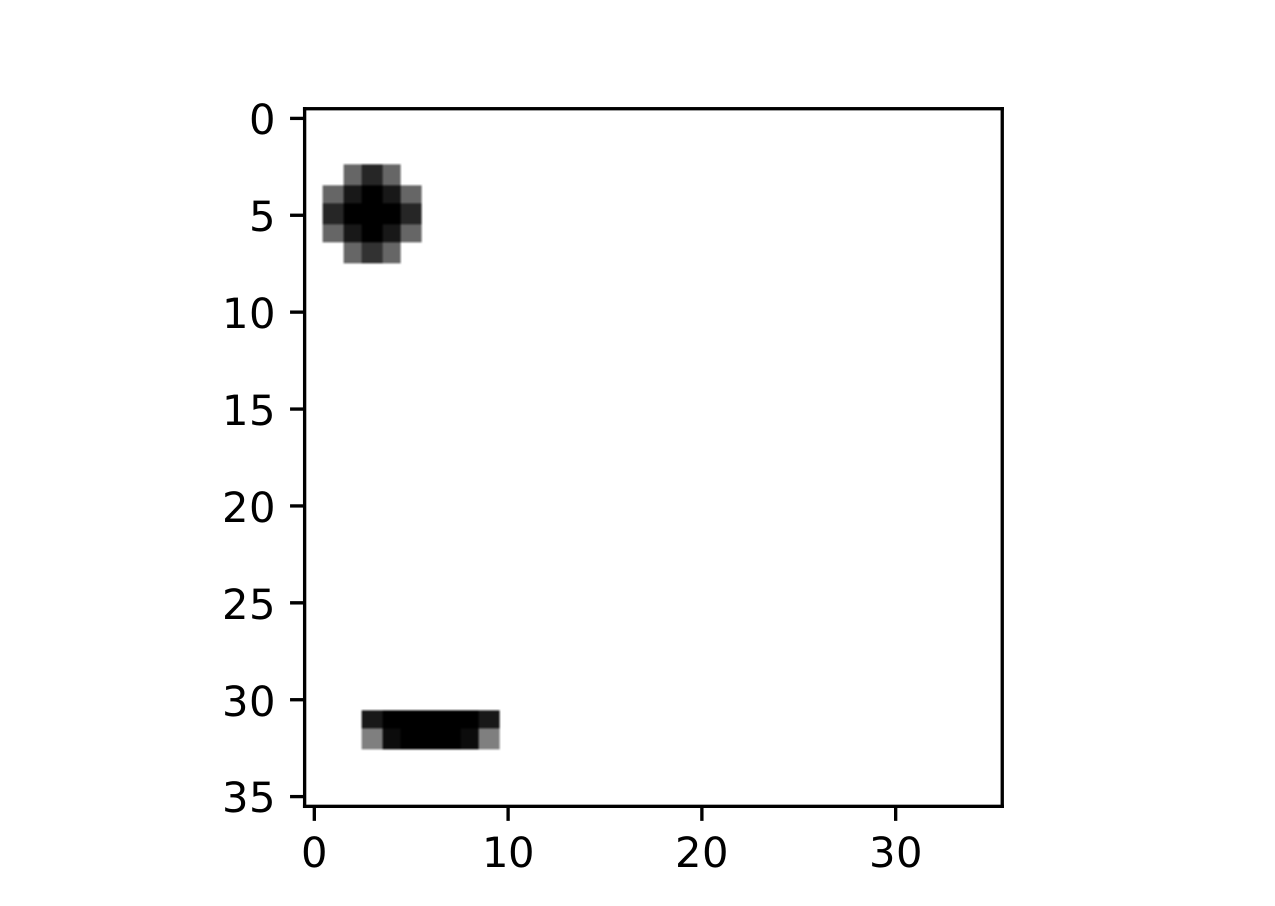}
  \caption{Representation of one state for the catcher environment.}
  \label{fig:pad_repr}
  \vspace{-10pt}
\end{wrapfigure}

The state representation is a two-dimensional array of $36 \times 36$ pixels $\in [-1,1]$.
This is illustrated in Figure~\ref{fig:pad_repr} and details are provided in Appendix~\ref{app:catcher}. 
This environment has only a few low-dimensional underlying important features for the state representation: (i)~the position of the paddle (one feature) and (ii)~the position of the blocks (two features). These features are sufficient to fully define the environment at any given time. This environment illustrates that the CRAR agent is not limited to navigation tasks and the difference with the previous example is that it has an actual reward function and model-free objective.

\begin{figure}[ht!]
  \centering
  \includegraphics[width=.6\linewidth]{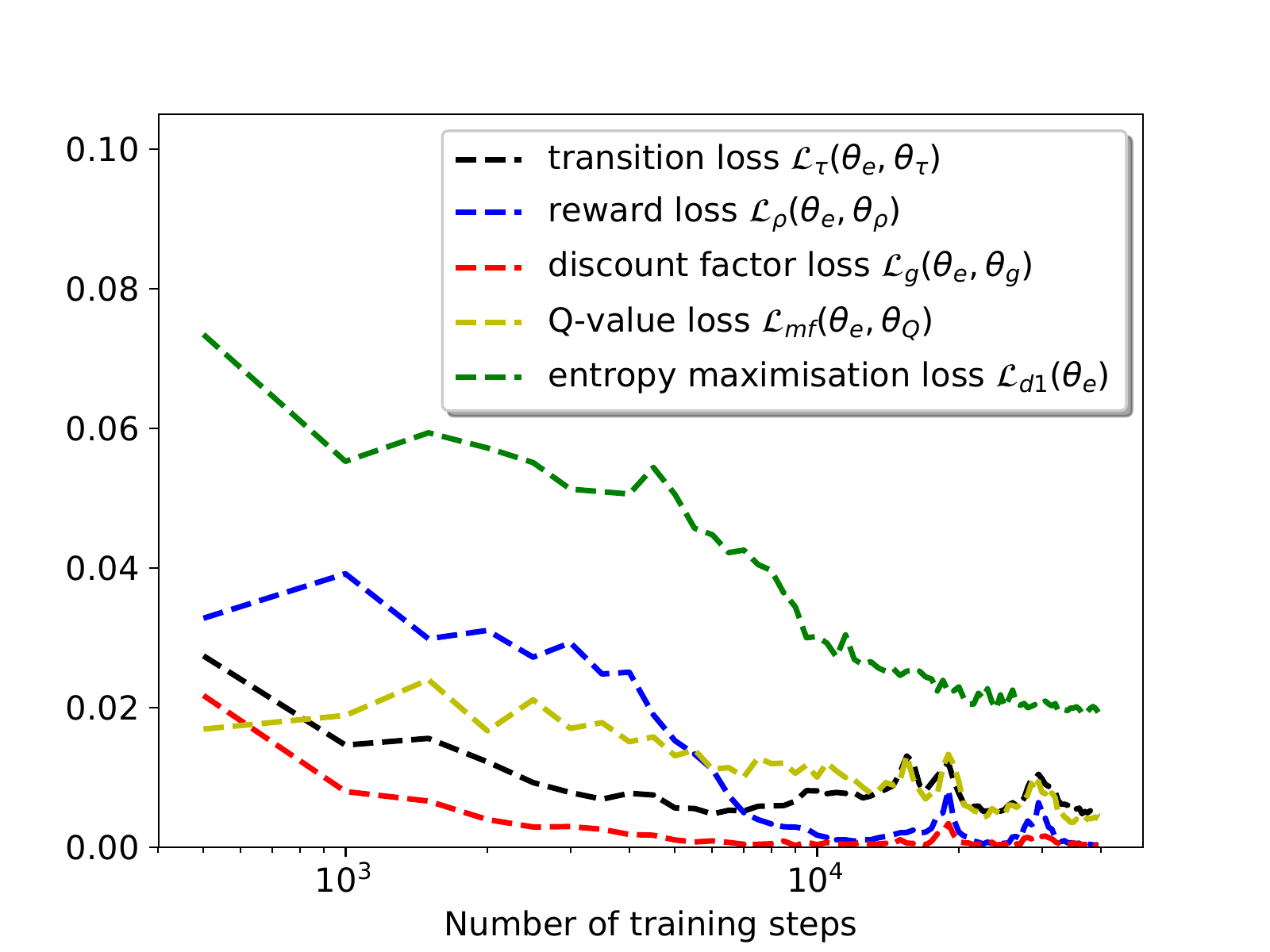}
  \caption{Representation of model-based and model-free losses through training in catcher. $\alpha= 5 \times10^{-4}$, $\beta=0.2$ and decreasing $\alpha$ by 10\% every 2000 training steps. All results obtained are qualitatively similar and robust to different learning rates as long as the initial learning rate $\alpha$ was not initialized to a too large value.}
  \label{fig:losses_repr_catcher}
\end{figure}

We show in Figure~\ref{fig:losses_repr_catcher} that all the losses behave well during training and that they can all together decrease to low values with a decreasing learning rate~$\alpha$. Note that all losses are learned through the abstract representation.

\begin{figure}[ht!]
\captionsetup[subfigure]{position=b}
\centering
\begin{subfigure}{.42\textwidth}
  \centering
  \includegraphics[width=1.0\linewidth]{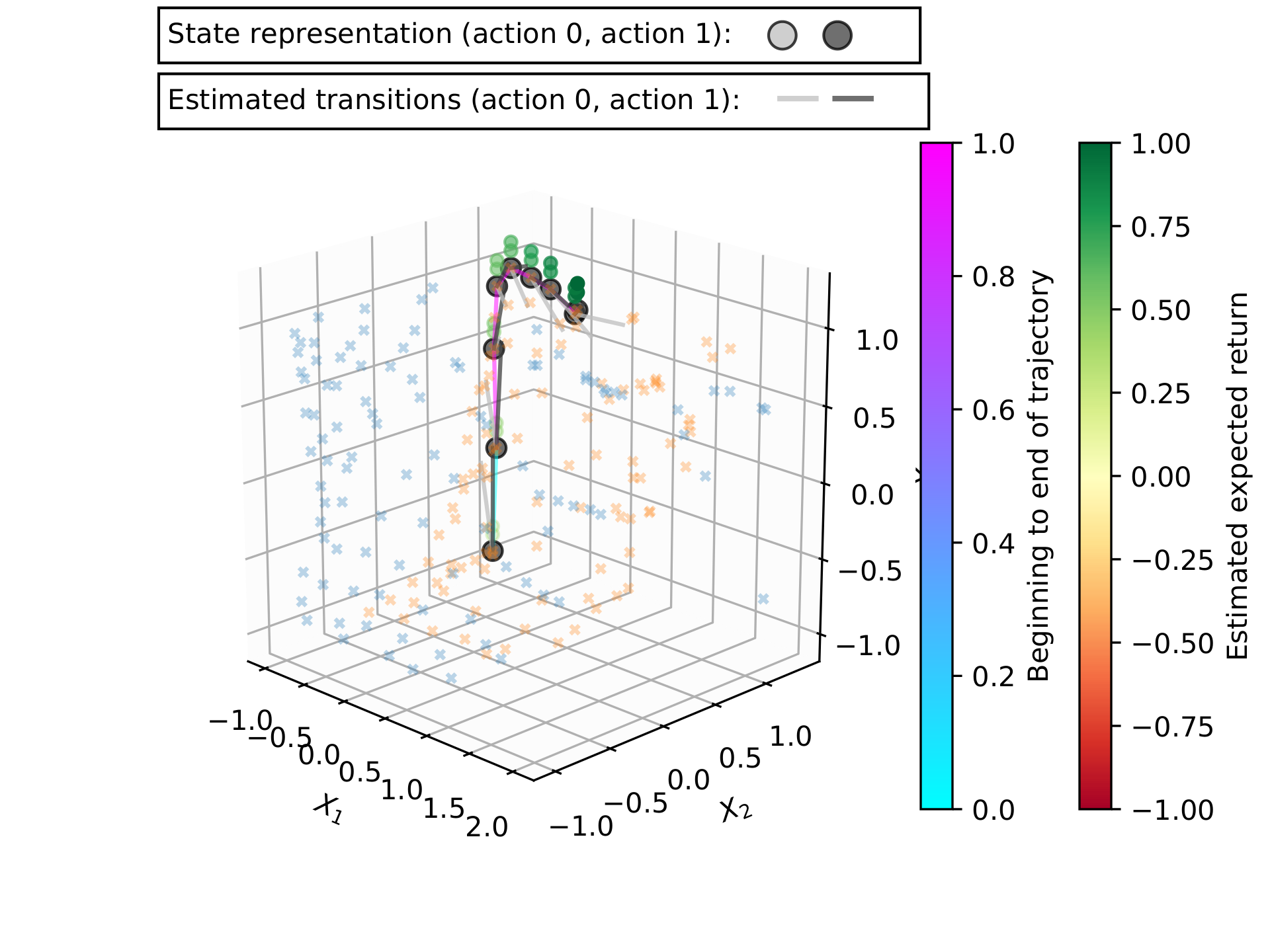}
   \caption{Without interpretability loss}
  \label{fig:pad_sub1}
\end{subfigure}%
\\ \vspace{0.2cm}
\begin{subfigure}{.42\textwidth}
  \centering
  \includegraphics[width=1.0\linewidth]{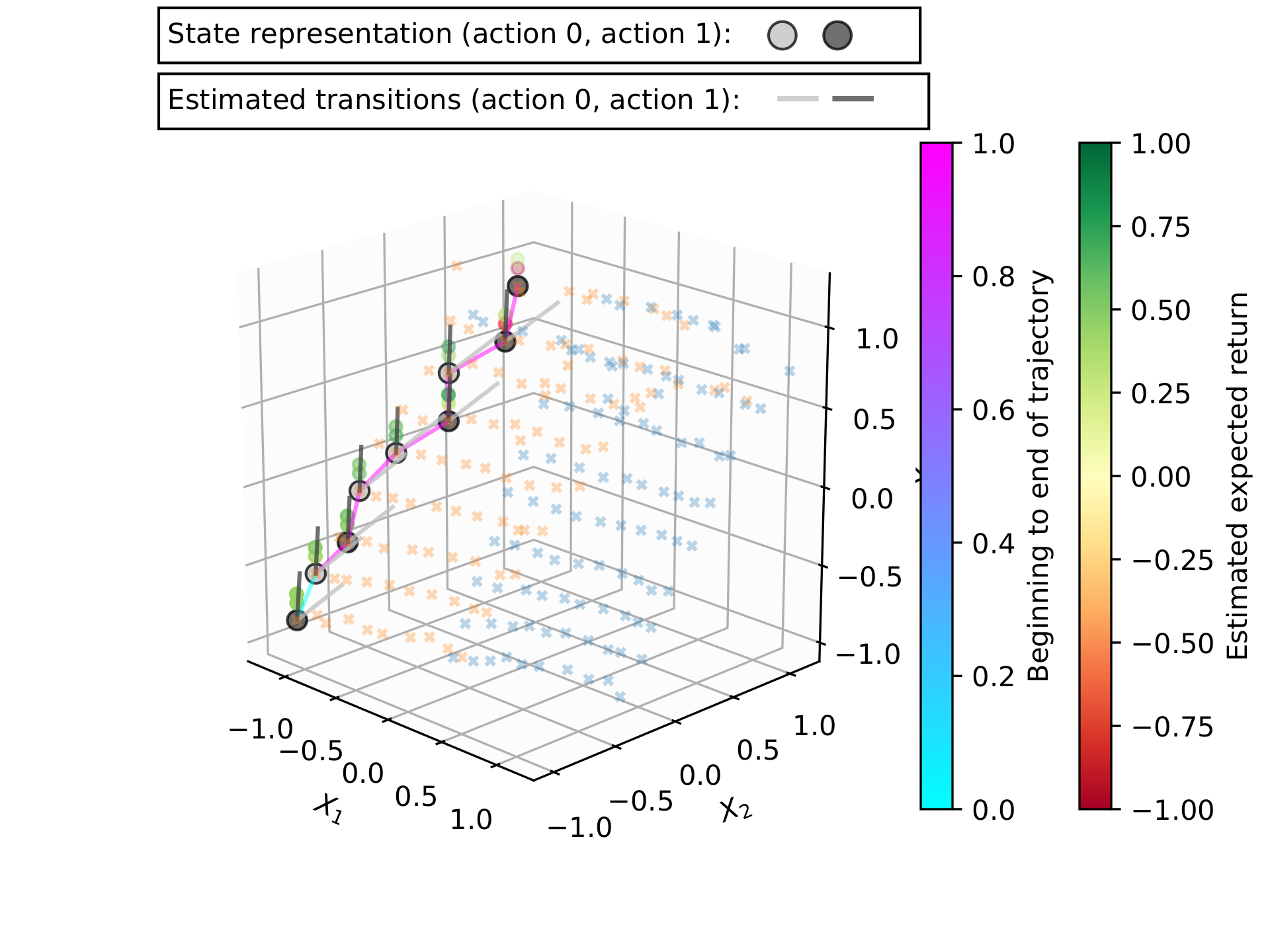}
  \caption{We use $v(a^{(1)})=(1,1)$ and $v(a^{(2)})=(-1,1)$ such that the first feature is forced to either increase or decrease depending on the action and the second feature is forced to increase with time (for both actions).}
  \label{fig:pad_sub2}
\end{subfigure}
\caption{Abstract representation of the domain by the CRAR agent after 50k training steps (details in the appendix). The blue and orange crosses represent respectively all possible reachable states for the ball starting respectively on the right and on the left. The trajectory is represented by the blue-purple curve (at the beginning a ball has just appeared). The colored dots represent the estimated expected return provided by $Q(x, a; \theta_Q )$. The actions taken are represented by the black/grey dots. The estimated transition are represented by straight lines (black for right, grey for left).}
\label{fig:pad}
\end{figure}

In Figure~\ref{fig:pad}, it is shown that the CRAR agent is able to build a three dimensional abstract representation of its environment.
Note that the CRAR agent is also able to catch the ball all the time (after 50k training steps and when following a greedy policy).


\subsection{Meta-learning with limited off-policy data}
\begin{wrapfigure}{R}{3.8cm}
\vspace{-0pt}
  \centering
  \includegraphics[width=.99\linewidth]{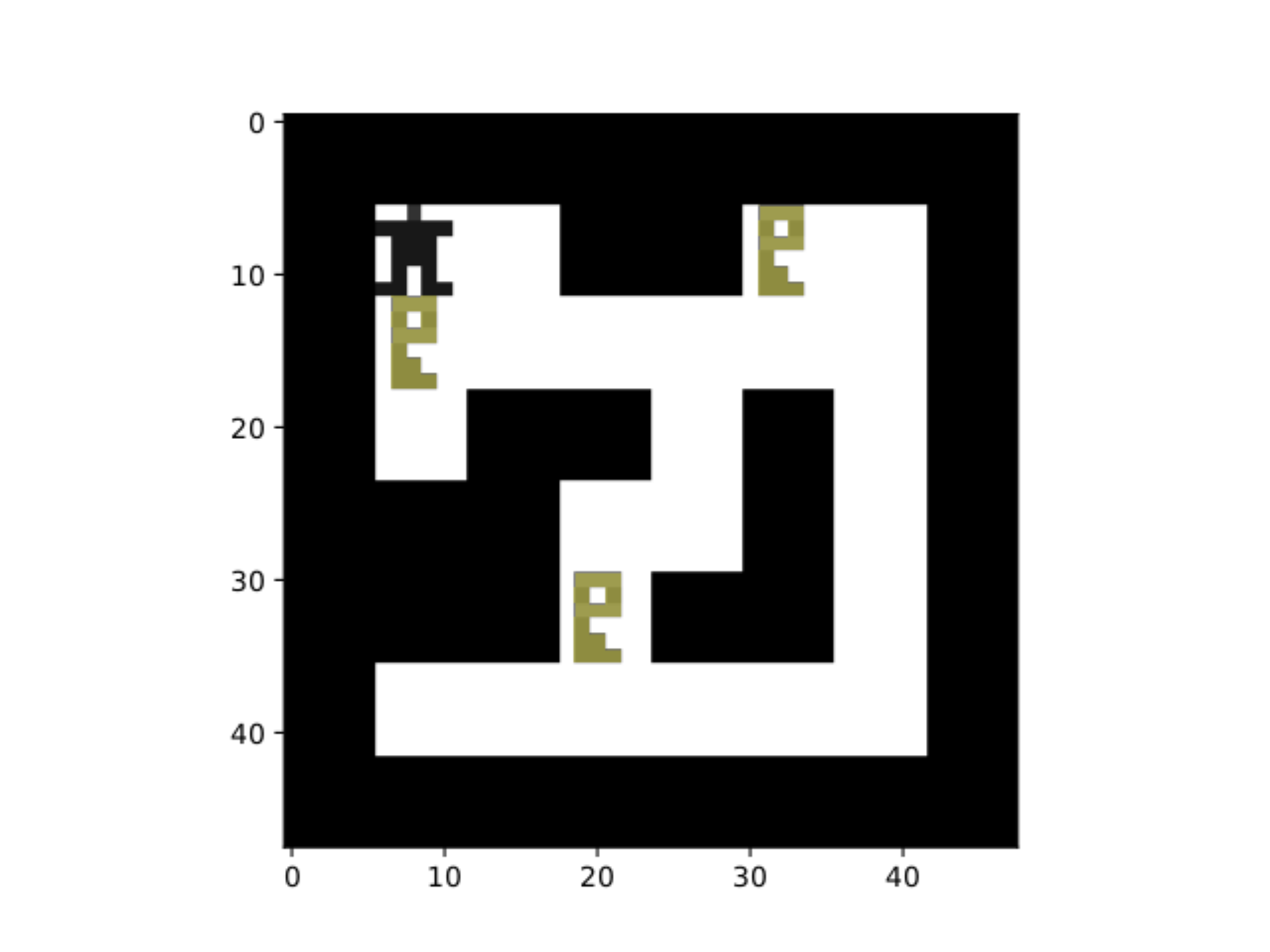}
  \caption{Representation of one state for one sample labyrinth with rewards.}
  \label{fig:lab_repr2}
\vspace{-5pt}
\end{wrapfigure}
The CRAR architecture can also be used in a meta-learning setting. We consider a distribution of labyrinth tasks (over reward locations, and wall configurations), where one sample is illustrated in Figure~\ref{fig:lab_repr2}. Overall, the empirical probability that two labyrinths taken randomly are the same is lower than $10^{-7}$; see details in the appendix.
The reward obtained by the agent is equal to $1$ when it reaches a key and it is equal to $-0.1$ for any other transition. 
We consider the batch RL setting where the agent has to build a policy offline from experience gathered following a purely random policy on a training set of $2 \times 10^5$ steps. This is equivalent to the set of transitions required in expectation to obtain the three keys by a random policy on about 500 different labyrinths (depending on the random seed). This setting makes up a more challenging task as compared to an online setting with (tens/hundreds of) millions of steps since the agent has to build a policy from a limited off-policy experience and requires strong generalization. In addition, it allows removing the exploration/exploitation influence from the experiment, thus easing the interpretation of the results.

\begin{figure}[ht!]
  \centering
  \includegraphics[width=.8\linewidth]{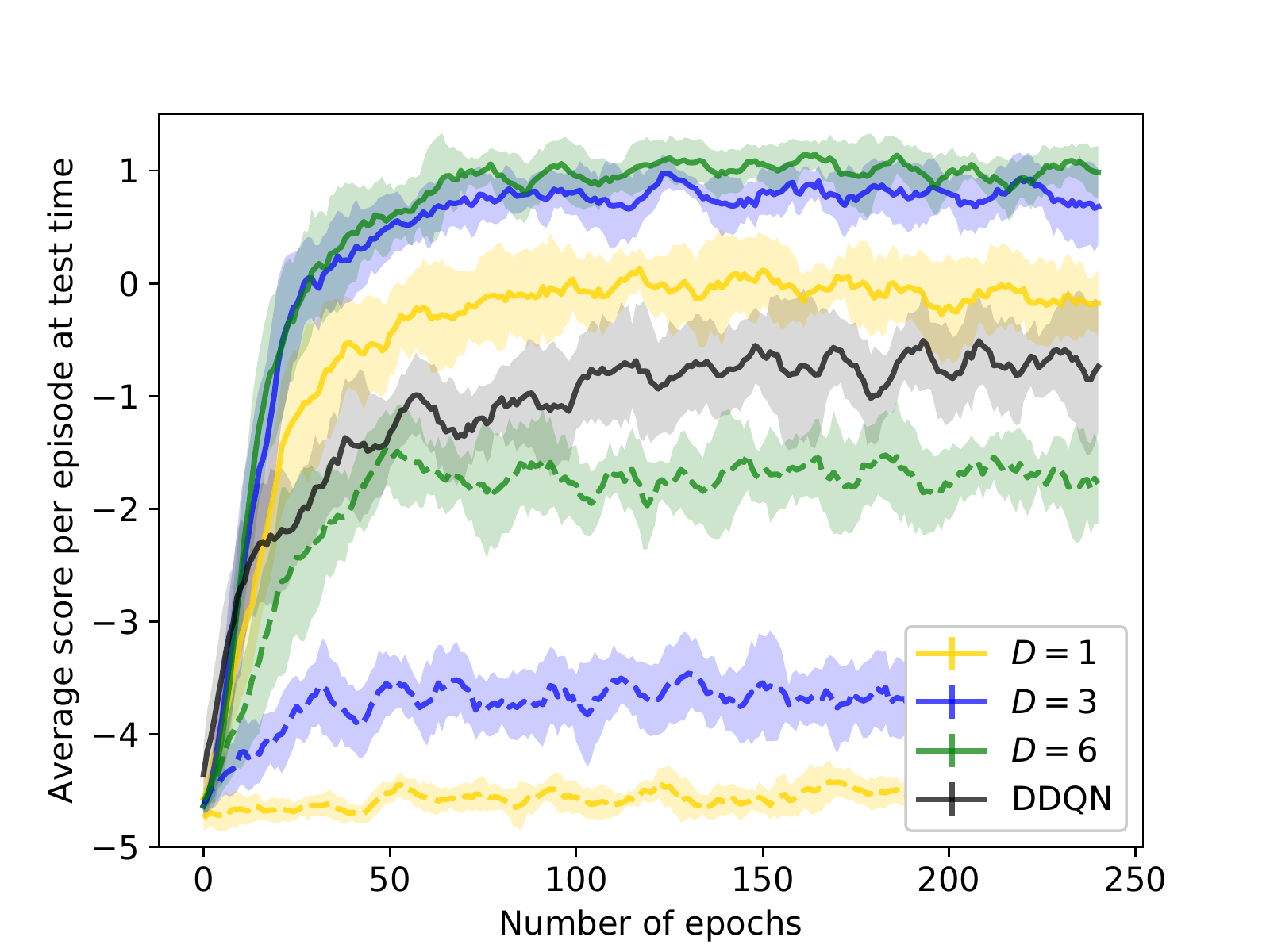}
\caption{Meta-learning score on a distribution of labyrinths where the training is done with a limited number of transitions obtained off-line by a random policy. An epoch is considered to be every $2000$ gradient descent steps (on all the losses). Every epoch, 200 steps on new labyrinths from the distribution are taken using different planning depths and the considered score is the running average of 10 such scores. The reported score is the mean of that running average along with the standard deviation (10 independent runs). Dotted lines represent policies without the model-free component.}
\label{fig:meta}
\end{figure}

In this context, we use a CRAR agent with an abstract state space made up of 3 channels, each of size $8 \times 8$ and the encoder is made up of CNNs only.
It is shown in Figure~\ref{fig:meta} that the CRAR agent is able to achieve better data efficiency by using planning (with depth $D=1,3,6$) as compared to pure model-free or pure model-based approaches.

The model-free DDQN baseline uses the same neural architecture but is trained only with the loss $\mathcal L_{mf}$.
As baselines, the pure model-based approach (represented as dotted lines) performs planning similarly to the CRAR agent, but selects the branches randomly and has a constant estimate of the value function at the leafs (when d=0 in Equation \ref{eq:Qd}). Note that, for a fair comparison, the model-based baselines have similar computational cost to take a decision than the CRAR agent for a given depth $d$; however, they have a worse performance due to the ablation of the model-free component. 

This experiment is, to the best of our knowledge, the first that is successfully able to learn efficiently from a small set of off-policy data in a complex distribution of tasks, while using planning in an abstract state space. A discussion of other similar works are provided in Section \ref{sec:mbmf}.

\subsection{Illustration of transfer learning}
\label{app:illu_transfer}
The CRAR architecture has the advantage of explicitly training its different components, and hence can be used for transfer learning by retraining/replacing some of its components to adjust to new tasks. In particular, one could enforce that states related to the same underlying task but with different renderings (e.g. real and simulation) are mapped into an abstract state that is close. In that case, an agent can be trained in simulation and then deployed in a realistic setting with limited retraining.

To illustrate the possibility of using the CRAR agent for transfer, we consider the setting where after 250 epochs, the high-dimensional state representations is now the negative of the previous representations\footnote{We define the negative of an image as the image where all pixels have the opposite value.}.
The experience available to the agent is the same as previously, except that all the (high-dimensional) state representations in the replay memory are converted to the negative images.
The transfer procedure consists in forcing, by supervised learning (with a MSE error and a learning rate of $5 \times 10^{-4}$), the encoder to fit the same abstract representation for 100 negative images than for the positive images (80 images are used as training set and 20 are used as validation set). 

\begin{figure}[ht!]
\captionsetup[subfigure]{position=b}
\centering
\begin{subfigure}{.45\textwidth}
  \centering
  \includegraphics[width=.72\linewidth]{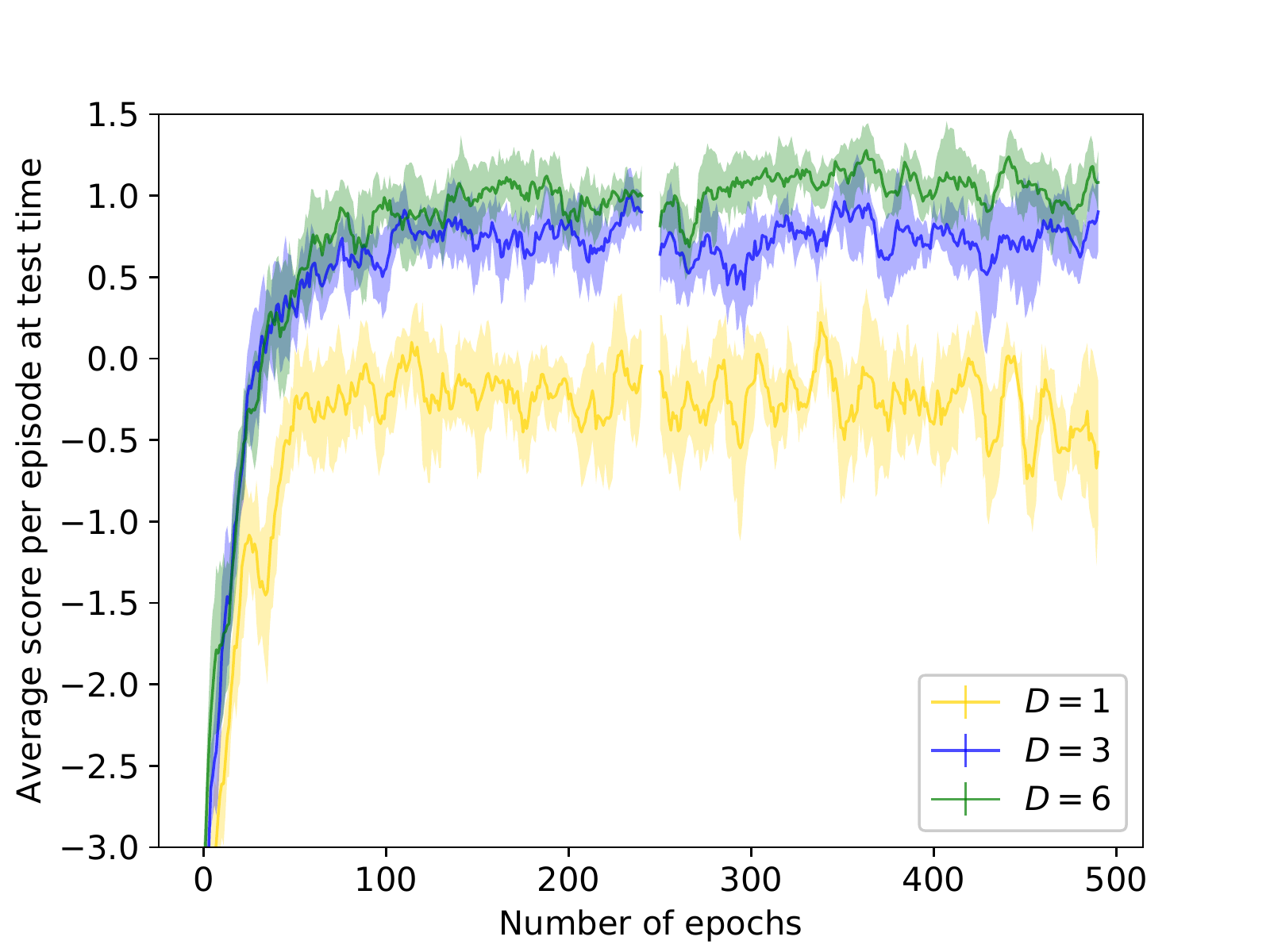}
  \caption{with transfer procedure at epoch 250.}
  \label{fig:meta_trans}
\end{subfigure}
 \ \ \ \ 
 \begin{subfigure}{.45\textwidth}
  \centering
  \includegraphics[width=.72\linewidth]{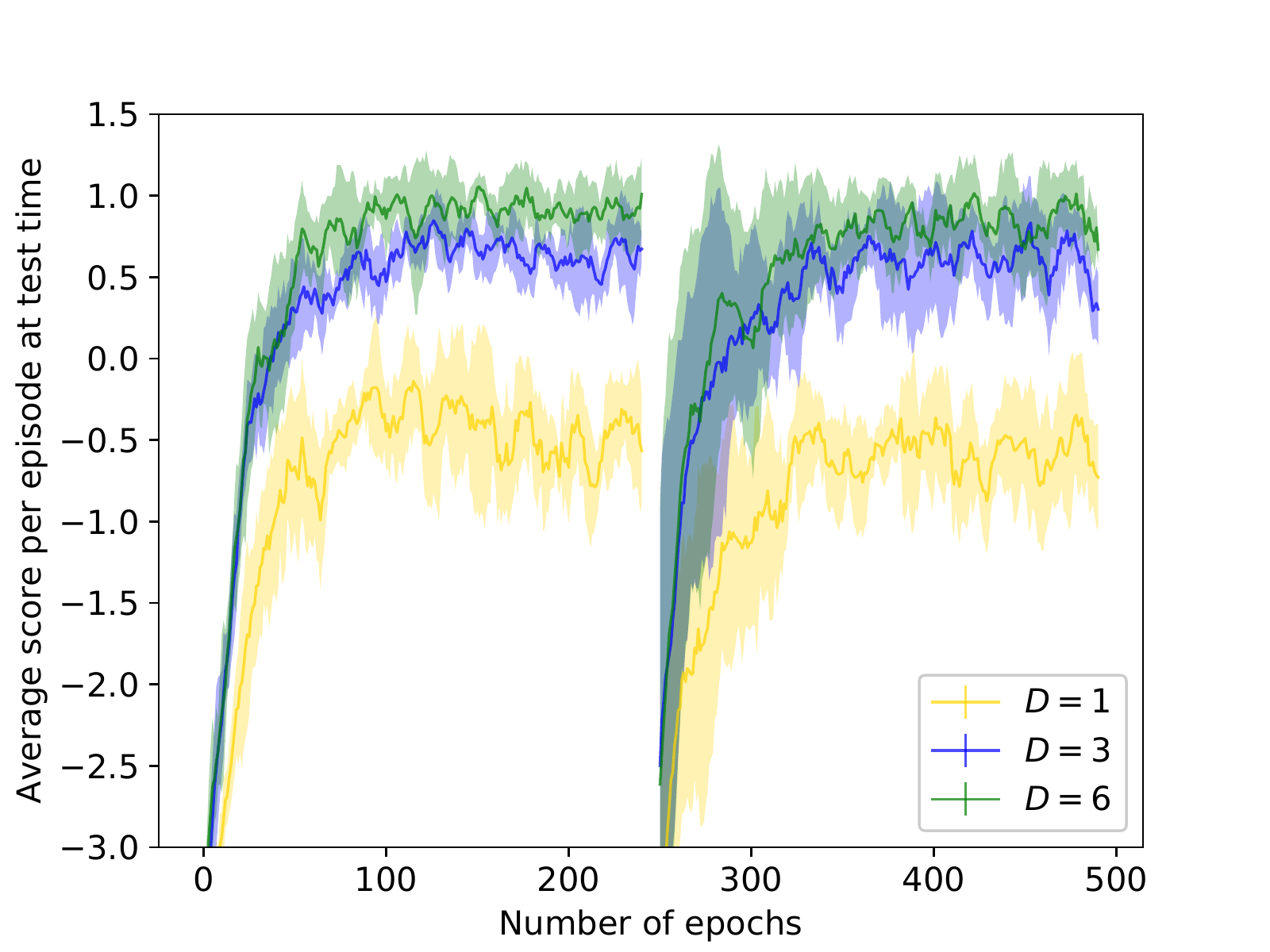}
  \caption{without transfer procedure at epoch 250.}
  \label{fig:meta_no_trans}
\end{subfigure}
\caption{The reported score is the mean of the running average along with the standard deviation (5 independent runs). On the first 250 epochs, training and test are done on the original distribution of labyrinths while for the remaining 250 epochs, training and test are done on the same distribution of tasks but where the states are negative images as compared to the original labyrinths.}
\label{fig:meta2}
\end{figure}

It can be seen in Figure~\ref{fig:meta_trans} that, with the transfer procedure, no retraining is necessary in contrast to Figure~\ref{fig:meta_no_trans}. Several other approaches exist to achieve this type of transfer; but this experiment demonstrates the flexibility of replacing some of the components of the CRAR agent to achieve transfer.

\section{Related work}
\label{sec:related}
\subsection{Building an abstract representation}
The idea of building an abstract representation with a low-dimensional representation of the important features for the task at hand 
is key in the whole field of deep learning and also highly prevalent in reinforcement learning. 
One of the key advantages of using a small but rich abstract representation is to allow for improved generalization. 

One approach is to first infer a factorized set of generative factors from the observations (e.g., with an encoder-decoder architecture variant \cite{zhang2018decoupling}, variational auto-encoder \cite{higgins2017darla} or using for instance t-SNE \cite{maaten2008visualizing}).
Then these features can be used as input to a reinforcement learning algorithm. The learned representation can, in some contexts, greatly help for generalization as it provides a more succinct representation that is less prone to overfitting.
In our setting, using an auto-encoder loss instead of the representation loss $\mathcal L_{d}$ does not ensure a sufficient diversity in the low-dimensional abstract space. This problem is illustrated in Appendix \ref{app:abla}.
In addition, an auto-encoder is often too strong of a constraint. On the one hand, some features may be kept in the abstract representation because they are important for the reconstruction of the observations, while they are otherwise irrelevant for the task at hand (e.g., the color of the cars in a self-driving car context). On the other hand, crucial information about the scene may also be discarded in the latent representation, particularly if that information takes up a small proportion of the observations $x$ in pixel space \cite{higgins2017darla}.


Another approach to build a set of relevant features is to share a common representation for solving a set of tasks.
The reason is that learning related tasks introduce an inductive bias that causes a model to build low level features in the neural network that can be useful for the range of tasks \cite{jaderberg2016reinforcement}.

The idea of an abstract representation can be found in neuroscience where the phenomenon of access consciousness can be seen as the formation of a low-dimensional combination of a few concepts which condition planning, communication and the interpretation of upcoming observations. In this context, the abstract state could be formed using an attention mechanism able to select specific relevant variables in a context-dependent manner \cite{bengio2017consciousness}.

In this work, we focus on building an abstract state that provides sufficient information to simultaneously fit an internal meaningful dynamics as well as the estimation of the expected value of an optimal policy.
The CRAR agent does not make use of any reconstruction loss, but instead learns both the model-free and model-based components through the state representation. By learning these components along with an approximate entropy maximization penalty, we have shown that the CRAR agent ensures that the low-dimensional representation of the task is meaningful.

\subsection{Integrating model-free and model-based}
\label{sec:mbmf}
Several recent works incorporate model-based and model-free RL and achieve improved sample efficiency.
The closest works to CRAR include the value iteration network (VIN)~\cite{tamar2016value}, the predictron~\cite{silver2016predictron} and the value prediction network (VPN) architecture~\cite{oh2017value}.
VIN is a fully differentiable neural network with a planning module that learns to plan from model-free objectives. 
As compared to CRAR, VIN has only been shown to work for navigation tasks from one initial position to one goal position. In addition, it does not work in a smaller abstract state space.
The predictron is aimed at developing an algorithm that is effective in the context of planning.
It works by \textit{implicitly} learning an internal model in an abstract state space which is used for policy evaluation.
The predictron is trained end-to-end to learn, from the abstract state space, (i)~the immediate reward and (ii)~value functions over multiple planning depths. 
The initial predictron architecture was limited to policy evaluation; it was then extended to learn an optimal policy through the VPN model. Since VPN relies on n-step Q-learning, it can not directly make use of off-policy data and is limited to the online setting.
As compared to these works, we show how it is possible to \textit{explicitly} learn both the model and a value function from off-policy data while ensuring that they are based on a shared sufficient state representation. In addition, our algorithm ensures a disentanglement of the low-dimensional abstract features, which opens up many possibilities. In particular, the obtained low-dimensional representation is still effective even in the absence of any reward (thus without the model-free part).

As compared to~\cite{kansky2017schema}, our approach relies directly on raw features (e.g. raw images) instead of an input of entity states. As compared to I2A~\cite{weber2017imagination} and many model-based approaches, 
our approach allows to build a model in an abstract low-dimensional space. This is more computationally efficient because planning can happen in the low-dimensional abstract state space. As compared to treeQn~\cite{farquhar2017treeqn}, the learning of the model is explicit and we show how that approach allows recovering a sufficient interpretable low-dimensional representation of the environment, even in the absence of model-free objectives.


\section{Discussion}
\label{sec:discussion}
In this paper, we have shown that it is possible to learn an abstract state representation thanks to both the model-free and model-based components as well as the approximate entropy maximization penalty. In addition, we have shown that the logical steps that require planning and estimating the expected return can happen in that low-dimensional abstract state space. 

Our architecture could be extended to the case of stochastic environments. For the model-based component, one could use a generative model conditioned on both the abstract state and the action (e.g., using a GAN~\cite{goodfellow2014generative}) and the planning algorithm should take into account the stochastic nature of the dynamics. Concerning the model-free components, it would be possible to use the distributional representation of the value function~\cite{bellemare2017distributional}.

\begin{figure}[ht!]
  \centering
  \includegraphics[width=0.65\linewidth]{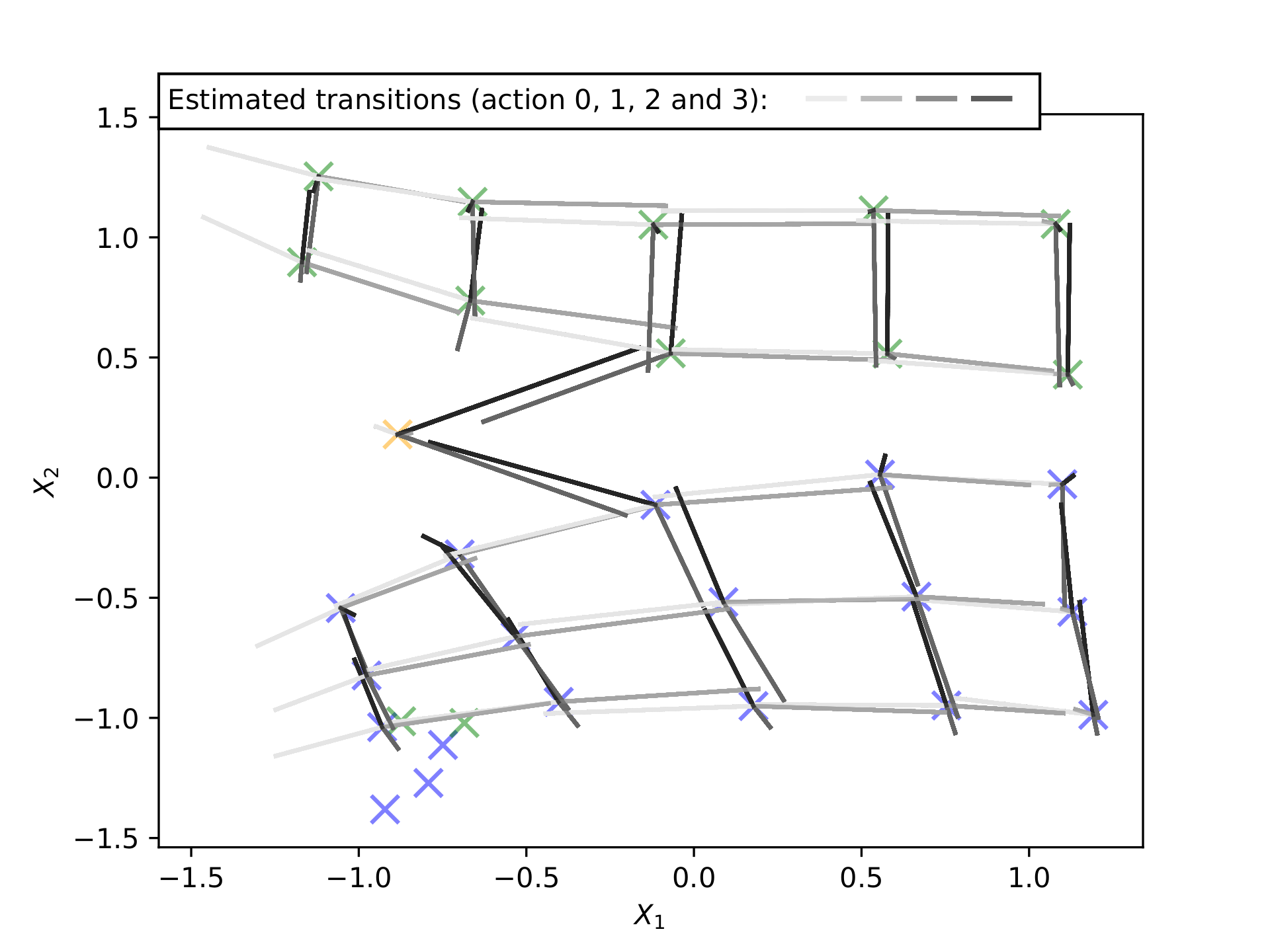}
  \caption{Abstract representation of the domain when the top part has not been explored (corresponding to the left part in this 2D low-dimensional representation). 
Thanks to the extrapolation abilities of the internal transition function, the CRAR agent can find a sequence of actions such that the expected representation of the new state is as far from any known abstract state (previously observed) for a given metric (e.g., $L_2$ distance).  Details related to this experiment are given in Appendix \ref{app:laby}.}
  \label{fig:lab_explo}
\end{figure}

Exploration is also one of the most important open challenges in deep reinforcement learning. 
The approach developed in this paper can be used as the basis for new exploration strategies as the CRAR agent provides a low-dimensional representation of the states, which can be used to more efficiently assess novelty.
An illustration of this is shown in Figure~\ref{fig:lab_explo}, in the context of the labyrinth task described in Section \ref{sec:experiments}.

Finally, in this paper, we have only considered a transition model for one time step.
An interesting future direction of work would be to incorporate temporal abstractions such as options, see e.g.~\cite{bacon2017option}.

\bibliographystyle{aaai}
\bibliography{references}{} 

\begin{thebibliography}{}

\bibitem[\protect\citeauthoryear{Abadi \bgroup et al\mbox.\egroup
  }{2016}]{abadi2016tensorflow}
Abadi, M.; Agarwal, A.; Barham, P.; Brevdo, E.; Chen, Z.; Citro, C.; Corrado,
  G.~S.; Davis, A.; Dean, J.; Devin, M.; et~al.
\newblock 2016.
\newblock Tensorflow: Large-scale machine learning on heterogeneous distributed
  systems.
\newblock {\em arXiv preprint arXiv:1603.04467}.

\bibitem[\protect\citeauthoryear{Bacon, Harb, and
  Precup}{2017}]{bacon2017option}
Bacon, P.-L.; Harb, J.; and Precup, D.
\newblock 2017.
\newblock The option-critic architecture.
\newblock In {\em AAAI},  1726--1734.

\bibitem[\protect\citeauthoryear{Belghazi \bgroup et al\mbox.\egroup
  }{2018}]{belghazi2018mine}
Belghazi, I.; Rajeswar, S.; Baratin, A.; Hjelm, R.~D.; and Courville, A.
\newblock 2018.
\newblock {MINE}: Mutual information neural estimation.
\newblock {\em arXiv preprint arXiv:1801.04062}.

\bibitem[\protect\citeauthoryear{Bellemare, Dabney, and
  Munos}{2017}]{bellemare2017distributional}
Bellemare, M.~G.; Dabney, W.; and Munos, R.
\newblock 2017.
\newblock A distributional perspective on reinforcement learning.
\newblock In {\em International Conference on Machine Learning},  449--458.

\bibitem[\protect\citeauthoryear{Bellman}{1957}]{bellman1957markovian}
Bellman, R.
\newblock 1957.
\newblock A markovian decision process.
\newblock {\em Journal of Mathematics and Mechanics}  679--684.

\bibitem[\protect\citeauthoryear{Bengio}{2017}]{bengio2017consciousness}
Bengio, Y.
\newblock 2017.
\newblock The consciousness prior.
\newblock {\em arXiv preprint arXiv:1709.08568}.

\bibitem[\protect\citeauthoryear{Chollet}{2015}]{chollet2015keras}
Chollet, F.
\newblock 2015.
\newblock Keras.
\newblock \url{https://github.com/fchollet/keras}.

\bibitem[\protect\citeauthoryear{Farquhar \bgroup et al\mbox.\egroup
  }{2017}]{farquhar2017treeqn}
Farquhar, G.; Rockt{\"a}schel, T.; Igl, M.; and Whiteson, S.
\newblock 2017.
\newblock Treeqn and atreec: Differentiable tree planning for deep
  reinforcement learning.
\newblock {\em arXiv preprint arXiv:1710.11417}.

\bibitem[\protect\citeauthoryear{Goodfellow \bgroup et al\mbox.\egroup
  }{2014}]{goodfellow2014generative}
Goodfellow, I.; Pouget-Abadie, J.; Mirza, M.; Xu, B.; Warde-Farley, D.; Ozair,
  S.; Courville, A.; and Bengio, Y.
\newblock 2014.
\newblock Generative adversarial nets.
\newblock In {\em Advances in neural information processing systems},
  2672--2680.

\bibitem[\protect\citeauthoryear{Hessel \bgroup et al\mbox.\egroup
  }{2017}]{hessel2017rainbow}
Hessel, M.; Modayil, J.; van Hasselt, H.; Schaul, T.; Ostrovski, G.; Dabney,
  W.; Horgan, D.; Piot, B.; Azar, M.; and Silver, D.
\newblock 2017.
\newblock Rainbow: Combining improvements in deep reinforcement learning.
\newblock {\em arXiv preprint arXiv:1710.02298}.

\bibitem[\protect\citeauthoryear{Higgins \bgroup et al\mbox.\egroup
  }{2017}]{higgins2017darla}
Higgins, I.; Pal, A.; Rusu, A.; Matthey, L.; Burgess, C.; Pritzel, A.;
  Botvinick, M.; Blundell, C.; and Lerchner, A.
\newblock 2017.
\newblock Darla: Improving zero-shot transfer in reinforcement learning.
\newblock In {\em International Conference on Machine Learning},  1480--1490.

\bibitem[\protect\citeauthoryear{Jaderberg \bgroup et al\mbox.\egroup
  }{2016}]{jaderberg2016reinforcement}
Jaderberg, M.; Mnih, V.; Czarnecki, W.~M.; Schaul, T.; Leibo, J.~Z.; Silver,
  D.; and Kavukcuoglu, K.
\newblock 2016.
\newblock Reinforcement learning with unsupervised auxiliary tasks.
\newblock {\em arXiv preprint arXiv:1611.05397}.

\bibitem[\protect\citeauthoryear{Kansky \bgroup et al\mbox.\egroup
  }{2017}]{kansky2017schema}
Kansky, K.; Silver, T.; M{\'e}ly, D.~A.; Eldawy, M.; L{\'a}zaro-Gredilla, M.;
  Lou, X.; Dorfman, N.; Sidor, S.; Phoenix, S.; and George, D.
\newblock 2017.
\newblock Schema networks: Zero-shot transfer with a generative causal model of
  intuitive physics.
\newblock In {\em International Conference on Machine Learning},  1809--1818.

\bibitem[\protect\citeauthoryear{Maaten and
  Hinton}{2008}]{maaten2008visualizing}
Maaten, L. v.~d., and Hinton, G.
\newblock 2008.
\newblock Visualizing data using t-sne.
\newblock {\em Journal of machine learning research} 9(Nov):2579--2605.

\bibitem[\protect\citeauthoryear{Mnih \bgroup et al\mbox.\egroup
  }{2015}]{mnih2015human}
Mnih, V.; Kavukcuoglu, K.; Silver, D.; Rusu, A.~A.; Veness, J.; Bellemare,
  M.~G.; Graves, A.; Riedmiller, M.; Fidjeland, A.~K.; Ostrovski, G.; et~al.
\newblock 2015.
\newblock Human-level control through deep reinforcement learning.
\newblock {\em Nature} 518(7540):529--533.

\bibitem[\protect\citeauthoryear{Mnih \bgroup et al\mbox.\egroup
  }{2016}]{mnih2016asynchronous}
Mnih, V.; Badia, A.~P.; Mirza, M.; Graves, A.; Lillicrap, T.~P.; Harley, T.;
  Silver, D.; and Kavukcuoglu, K.
\newblock 2016.
\newblock Asynchronous methods for deep reinforcement learning.
\newblock In {\em International Conference on Machine Learning}.

\bibitem[\protect\citeauthoryear{Oh, Singh, and Lee}{2017}]{oh2017value}
Oh, J.; Singh, S.; and Lee, H.
\newblock 2017.
\newblock Value prediction network.
\newblock In {\em Advances in Neural Information Processing Systems},
  6120--6130.

\bibitem[\protect\citeauthoryear{Silver \bgroup et al\mbox.\egroup
  }{2016}]{silver2016predictron}
Silver, D.; van Hasselt, H.; Hessel, M.; Schaul, T.; Guez, A.; Harley, T.;
  Dulac-Arnold, G.; Reichert, D.; Rabinowitz, N.; Barreto, A.; et~al.
\newblock 2016.
\newblock The predictron: End-to-end learning and planning.
\newblock {\em arXiv preprint arXiv:1612.08810}.

\bibitem[\protect\citeauthoryear{Tamar \bgroup et al\mbox.\egroup
  }{2016}]{tamar2016value}
Tamar, A.; Levine, S.; Abbeel, P.; WU, Y.; and Thomas, G.
\newblock 2016.
\newblock Value iteration networks.
\newblock In {\em Advances in Neural Information Processing Systems},
  2146--2154.

\bibitem[\protect\citeauthoryear{van Hasselt, Guez, and
  Silver}{2016}]{van2016deep}
van Hasselt, H.; Guez, A.; and Silver, D.
\newblock 2016.
\newblock Deep reinforcement learning with double q-learning.
\newblock In {\em Thirtieth AAAI Conference on Artificial Intelligence}.

\bibitem[\protect\citeauthoryear{Weber \bgroup et al\mbox.\egroup
  }{2017}]{weber2017imagination}
Weber, T.; Racani{\`e}re, S.; Reichert, D.~P.; Buesing, L.; Guez, A.; Rezende,
  D.~J.; Badia, A.~P.; Vinyals, O.; Heess, N.; Li, Y.; et~al.
\newblock 2017.
\newblock Imagination-augmented agents for deep reinforcement learning.
\newblock {\em arXiv preprint arXiv:1707.06203}.

\bibitem[\protect\citeauthoryear{White}{2016}]{white2016unifying}
White, M.
\newblock 2016.
\newblock Unifying task specification in reinforcement learning.
\newblock {\em arXiv preprint arXiv:1609.01995}.

\bibitem[\protect\citeauthoryear{Zhang, Satija, and
  Pineau}{2018}]{zhang2018decoupling}
Zhang, A.; Satija, H.; and Pineau, J.
\newblock 2018.
\newblock Decoupling dynamics and reward for transfer learning.
\newblock {\em arXiv preprint arXiv:1804.10689}.

\end{thebibliography}

\newpage
\appendix
\section{Generic algorithm details used in all experiments}
\label{app:NN}

The architecture of the different elements are detailed hereafter. Except when stated otherwise, all the elements of the architecture described in this section have been used for all experiments. "Conv2D" refers to a 2D convolutional layer (the stride is 1 and the padding is such that the output layer has the same dimensions, except the number of channels). "MaxPooling2D" refers to a pooling operation. "Dense" refers to a fully connected layer. The algorithm has used Keras~\cite{chollet2015keras} and Tensorflow~\cite{abadi2016tensorflow}.
$C_d$ is a constant set to 5. The batch size is 32. The freeze interval for the target parameters $\theta_k^{-}$ is 1000 steps.

\subsection{Encoder}
The encoder is made up of the succession of the following layers:
\begin{itemize}
\item Conv2D (8 channels, (2, 2) kernel, activation='tanh'),
\item Conv2D(16 channels, (2, 2) kernel, activation='tanh'), 
\item MaxPooling2D(pool size (2, 2)),
\item Conv2D(32 channels, (3, 3) kernel, activation='tanh'), 
\item MaxPooling2D(pool size (3, 3)).
\end{itemize}

For the abstract state made up of $n_a$ unstructured neurons (e.g., $n_a=2$ or $3$), it is followed by
\begin{itemize}
\item Dense(200 neurons, activation='tanh'),
\item Dense(100 neurons, activation='tanh'),
\item Dense(50 neurons, activation='tanh'),
\item Dense(10 neurons, activation='tanh'),
\item Dense($n_a$ neurons).
\end{itemize}

For the abstract state made up of neurons that keeps locality information, it is followed by the following layer:
\begin{itemize}
\item Conv2D($n_c$, (1, 1)),
\end{itemize}
where $n_c$ is the number of internal channels.

\subsection{Transition model}
The concatenation of the abstract state and the action is provided as input.
For the unstructured abstract state, the transition model is made up by
\begin{itemize}
\item Dense(10 neurons, activation='tanh'),
\item Dense(30 neurons, activation='tanh'),
\item Dense(30 neurons, activation='tanh'),
\item Dense(10 neurons, activation='tanh'),
\item Dense($n_a$ neurons).
\end{itemize}

Otherwise, the transition model is made up by
\begin{itemize}
\item Conv2D(16 channels, (1, 1) kernel, activation='tanh'),
\item Conv2D(32 channels, (2, 2) kernel, activation='tanh'),
\item Conv2D(64 channels, (3, 3) kernel, activation='tanh'),
\item Conv2D(32 channels, (2, 2) kernel, activation='tanh'),
\item Conv2D(16 channels, (1, 1) kernel, activation='tanh').
\end{itemize}

\subsection{Reward and discount factor models}
The concatenation of the abstract state and the action is provided as input.
For the unstructured abstract state, the transition model is made up by
\begin{itemize}
\item Conv2D(16 channels, (2, 2) kernel, activation='tanh'),
\item Conv2D(32 channels, (3, 3) kernel, activation='tanh'),
\item MaxPooling2D(pool size (2, 2)),
\item Conv2D(16 channels, (2, 2) kernel, activation='tanh'),
\item Conv2D(4 channels, (1, 1) kernel, activation='tanh').
\item Dense(200, activation='tanh')(x),
\item Dense(50 neurons, activation='tanh'),
\item Dense(20 neurons, activation='tanh') ,     
\item Dense(1 neuron).
\end{itemize}
Otherwise:
\begin{itemize}
\item Dense(10 neurons, activation='tanh'),
\item Dense(50 neurons, activation='tanh'),
\item Dense(20 neurons, activation='tanh'),     
\item Dense(1 neuron).
\end{itemize}

\subsection{Q-network}     
The abstract state is provided as input.
For the structured abstract state, the transition model is made up by
\begin{itemize}
\item Conv2D(16 channels, (2, 2) kernel, activation='tanh'),
\item Conv2D(32 channels, (3, 3) kernel, activation='tanh'),
\item MaxPooling2D(pool size (2, 2))
\item Conv2D(16 channels, (2, 2) kernel, activation='tanh'),
\item Conv2D(4 channels, (1, 1) kernel, activation='tanh'),
\item Dense(200 neurons, activation='tanh'),
\item Dense(50 neurons, activation='tanh'),
\item Dense(20 neurons, activation='tanh'),
\item Dense(number of actions).
\end{itemize}

For the unstructured abstract state, the transition model is made up by
\begin{itemize}
\item Dense(20 neurons, activation='tanh')
\item Dense(50 neurons, activation='tanh')
\item Dense(20 neurons, activation='tanh')
\item Dense(number of actions)
\end{itemize}

\section{Details on the single labyrinth environment task}
\label{app:laby}
Even though the dynamics happens in a $8 \times 8$ grid, the state representation is a 2 dimensional array of $48 \times 48$ pixels $\in [-1,1]$. For Figures \ref{fig:lab_dis}, \ref{fig:lab_forced} and \ref{fig:lab_explo}, the dataset is made up of 5000 transitions obtained with a purely random policy in the domain (10 epoch of 500 transitions, where each epoch starts with the agent in the corridor). 
The following learning rates are used:
$\alpha=5 \times 10^{-4}$ (decreased by 10\% every 2000 training steps such that the losses converge close to zero);
$\beta=0.2$;
$\alpha_{interpr}=\alpha/2$ (when used).
All figures presented are with 100k training steps (except when stated otherwise).

For Figure~\ref{fig:lab_explo}, the dataset is also made up of 5000 transitions obtained with a purely random policy in the domain except that all transitions that lead the agent to transition to the top part are discarded and the agent starts a new epoch.

\subsection{Ablation study and sensitivity study to hyper-parameters}
\label{app:abla}

We conduct in the section a sensitivity analysis and an ablation study on the simple labyrinth experiment (starting from the same setting than in Figure~\ref{fig:lab_dis}).
 
\begin{figure}[ht!]
    \centering
    \begin{subfigure}[t]{0.22\textwidth}
        \centering
        \includegraphics[width=0.99\linewidth]{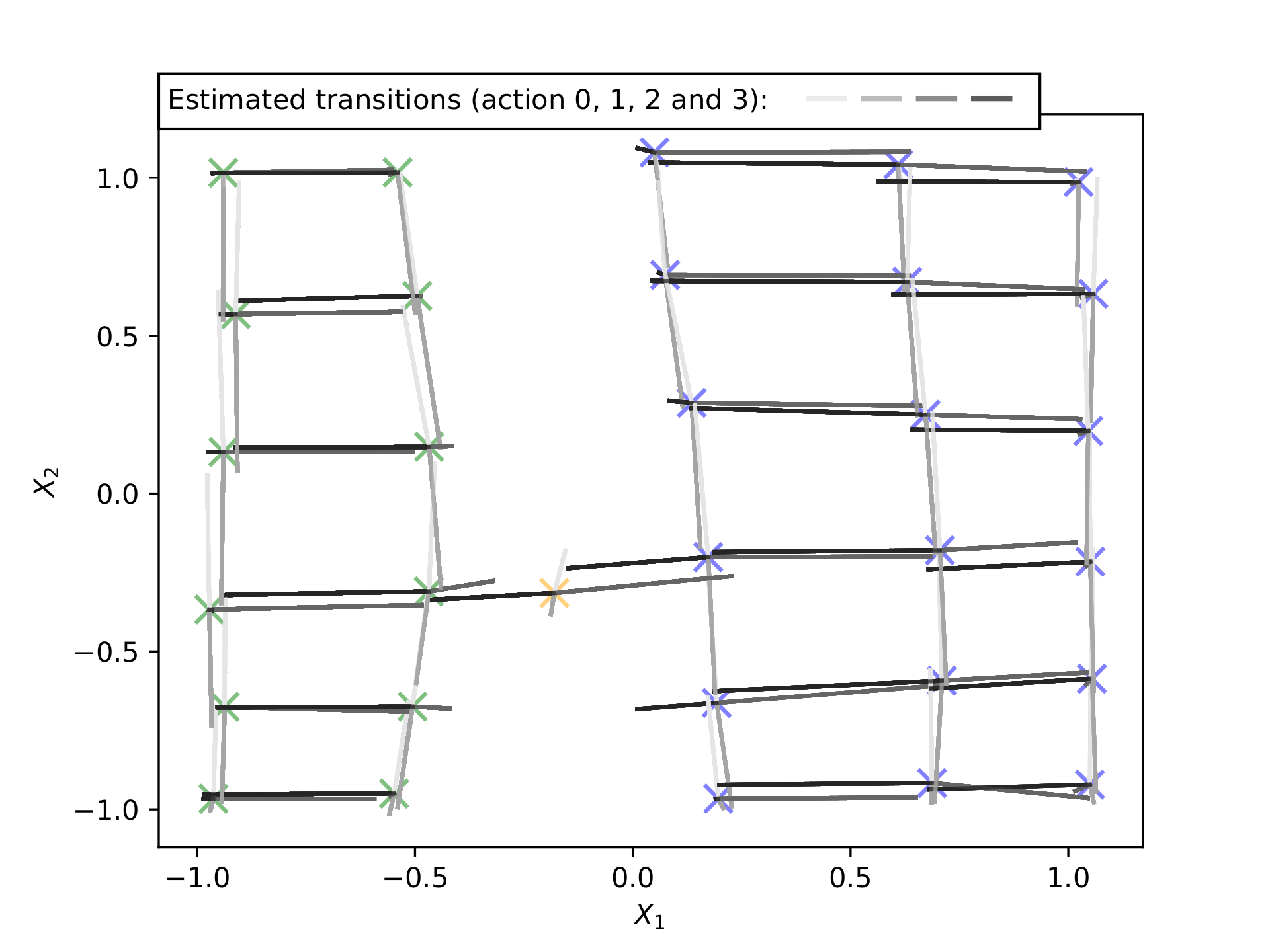}
        \caption{$\beta=0$.}
        \label{fig:lab_dis_abla1}
    \end{subfigure}%
    \ \ \
    \begin{subfigure}[t]{0.22\textwidth}
        \centering
        \includegraphics[width=0.99\linewidth]{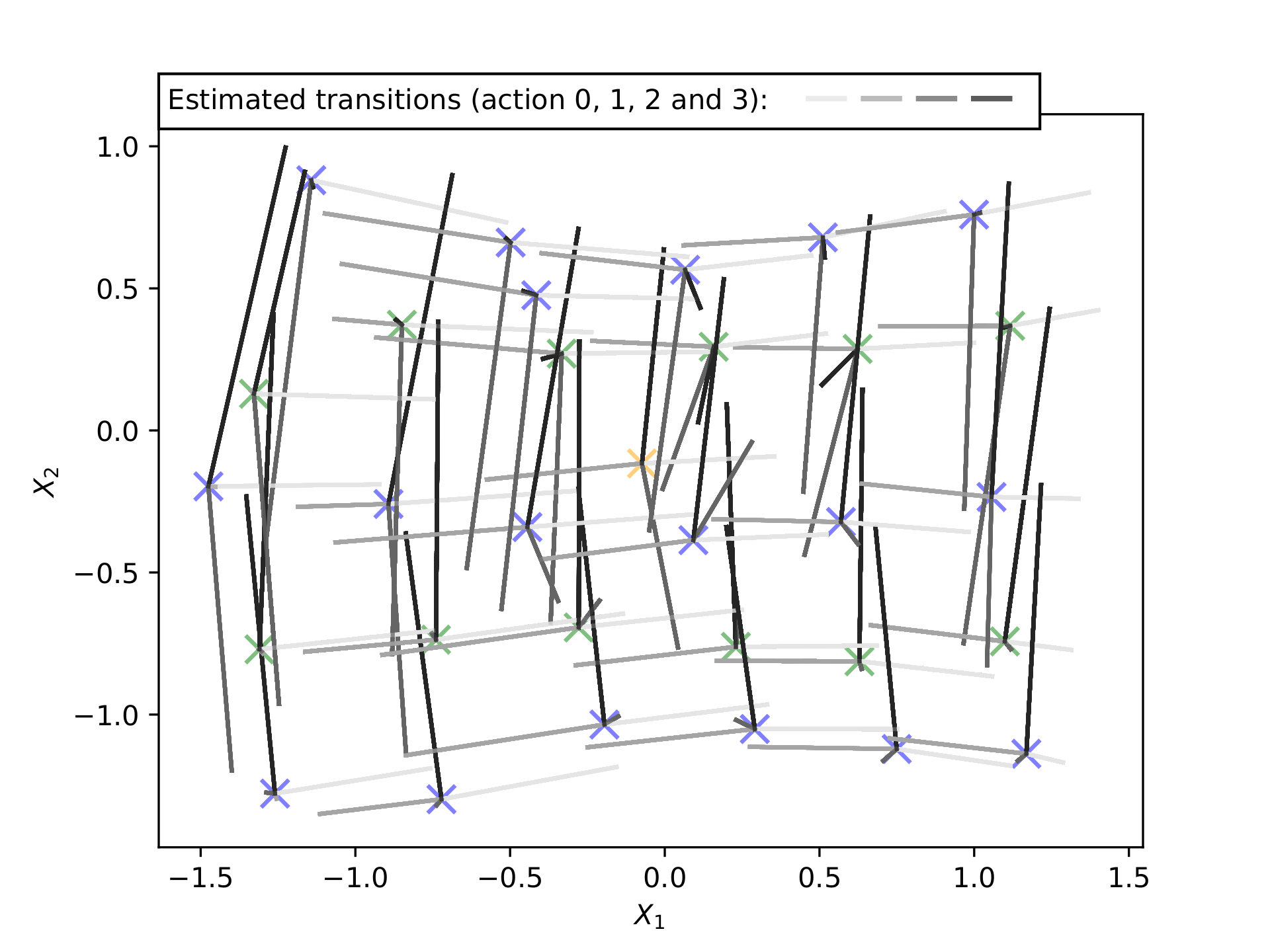}
        \caption{$\beta=0.5$.}
        \label{fig:lab_dis_abla2}
    \end{subfigure}
    \caption{Sensitivity study of the hyperparameter $\beta$.}
\end{figure}

The CRAR agent can capture a meaningful representation of the maze environment, even without any particular resampling of successive states for the entropy loss (see Figure~\ref{fig:lab_dis_abla1}). When a strong entropy between successive states is enforced,  Figure~\ref{fig:lab_dis_abla2} shows that this prevents a more natural representation. However, one can note that some meaningful structure is still preserved.

\begin{figure}[ht!]
    \centering
    \begin{subfigure}[t]{0.22\textwidth}
        \centering
        \includegraphics[width=0.99\linewidth]{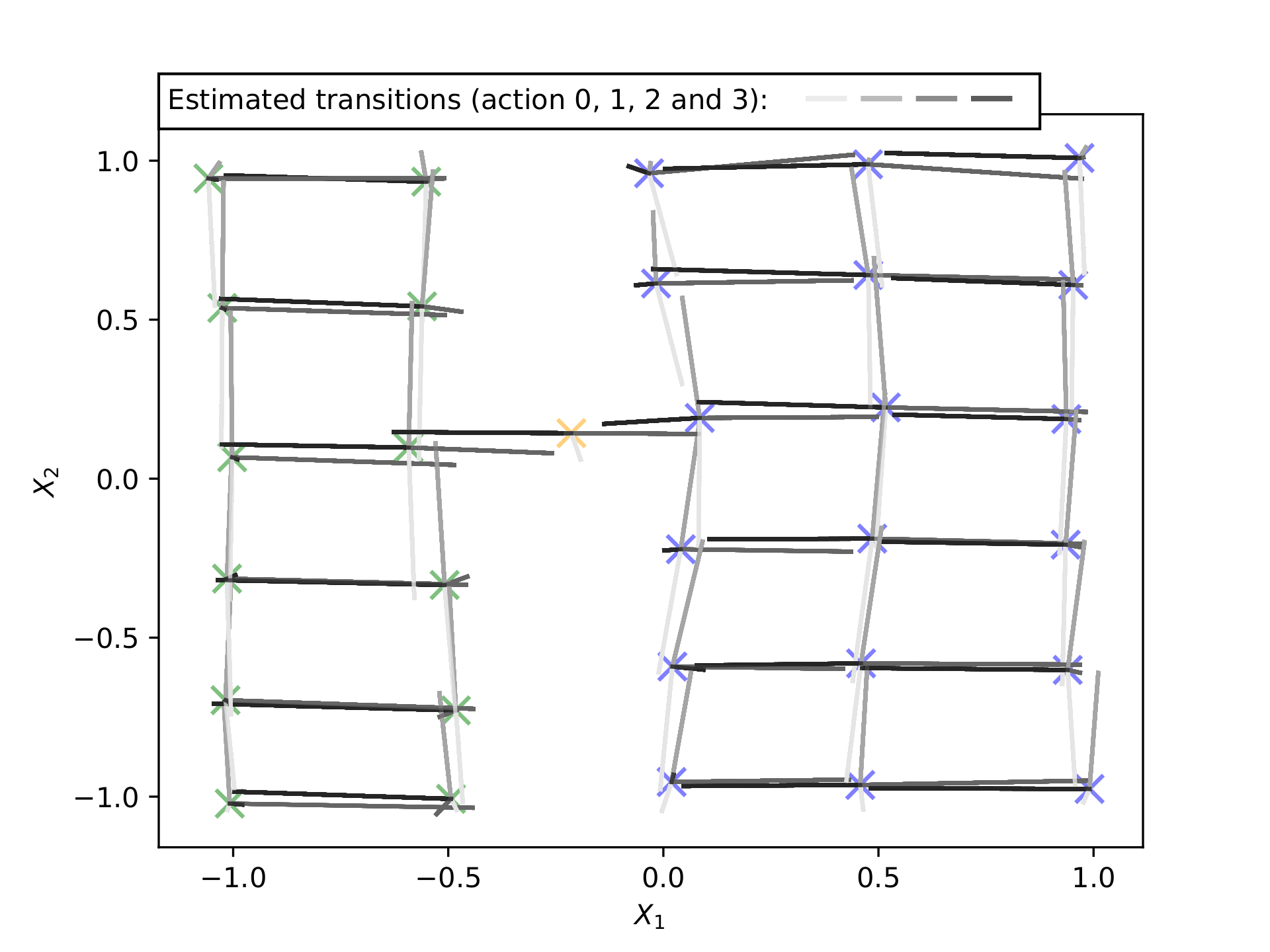}
        \caption{$\alpha$ fixed at $0.0001$.}
        \label{fig:lab_dis_abla3}
    \end{subfigure}%
    \ \ \
    \begin{subfigure}[t]{0.22\textwidth}
        \centering
        \includegraphics[width=0.99\linewidth]{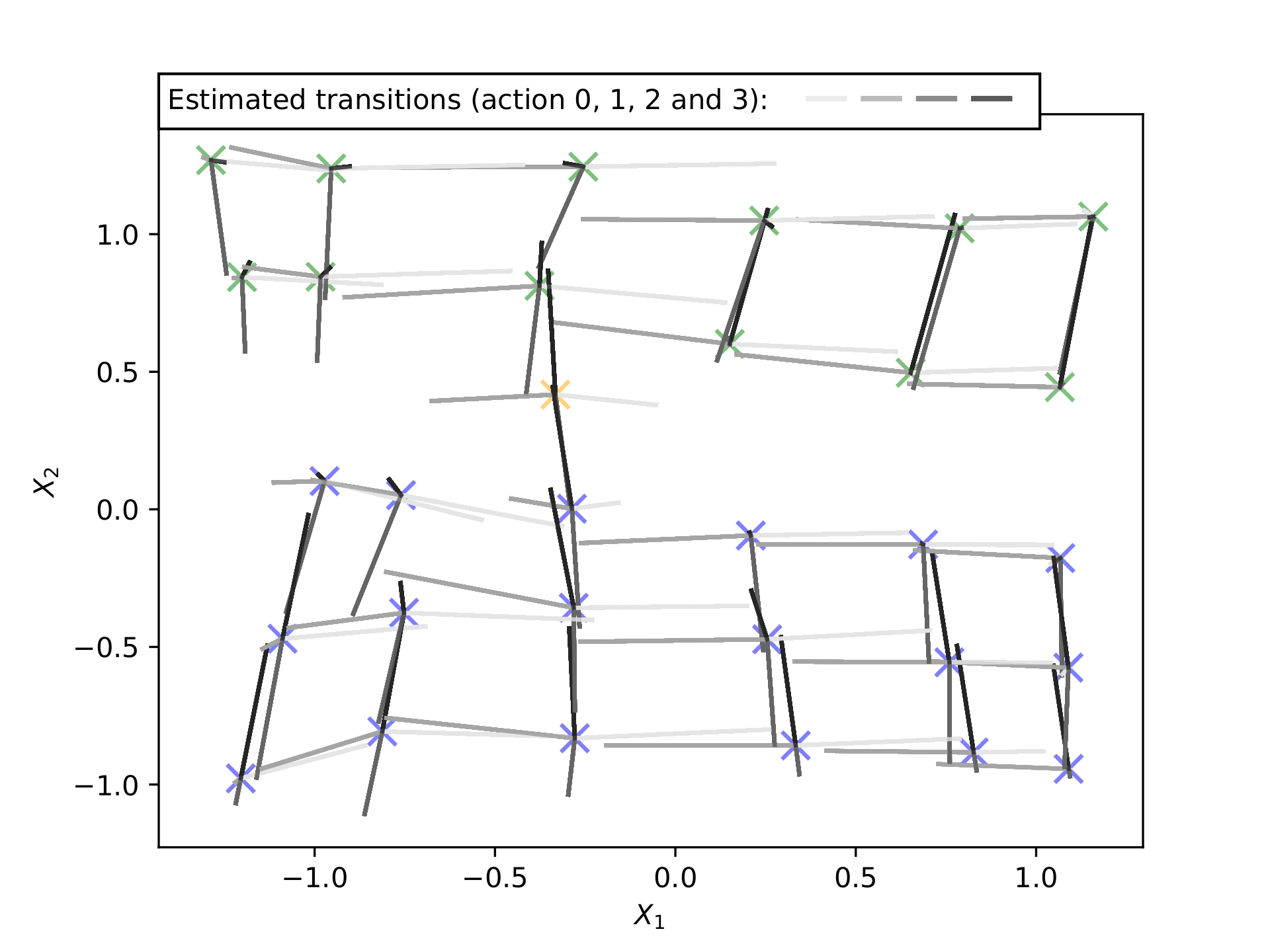}
        \caption{$\alpha$ fixed at $0.0005$}
        \label{fig:lab_dis_abla4}
    \end{subfigure}
    \caption{Sensitivity study of the learning rate $\alpha$.}
\end{figure}

As long as the learning rate is sufficiently small, the CRAR agent is able to built an accurate representation of the maze environment (see Figure~\ref{fig:lab_dis_abla3}). When the learning rate is too large, some instabilities prevent the transition function between abstract states to be accurate (see Figure~\ref{fig:lab_dis_abla4}).

\begin{figure}[ht!]
    \centering
    \begin{subfigure}[t]{0.22\textwidth}
  \centering
  \includegraphics[width=0.99\linewidth]{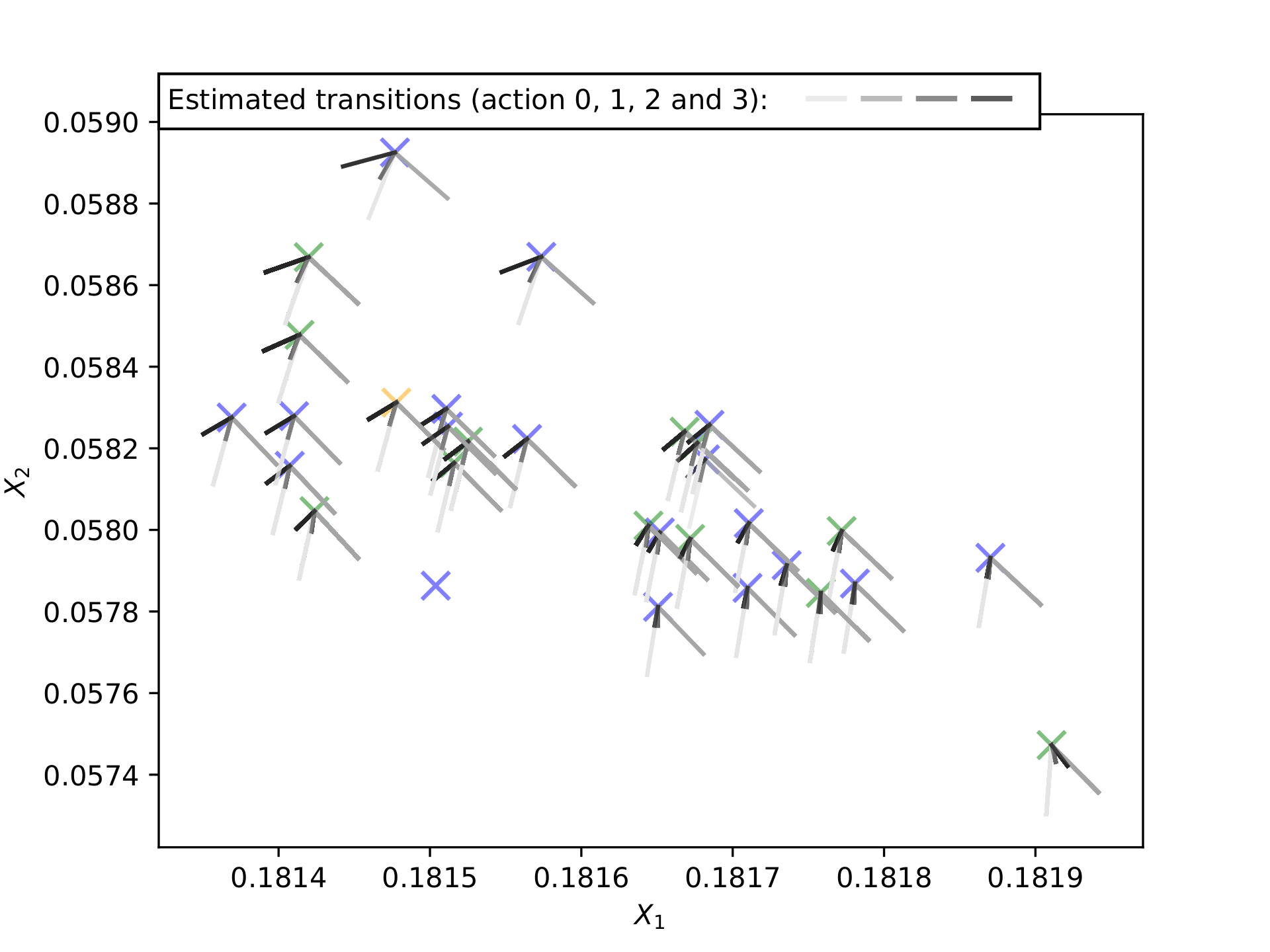}
  \caption{Ablation of the representation loss $\mathcal L_{d}$.}
  \label{fig:lab_dis_abla5}
    \end{subfigure}%
    \ \ \
    \begin{subfigure}[t]{0.22\textwidth}
  \centering
  \includegraphics[width=0.99\linewidth]{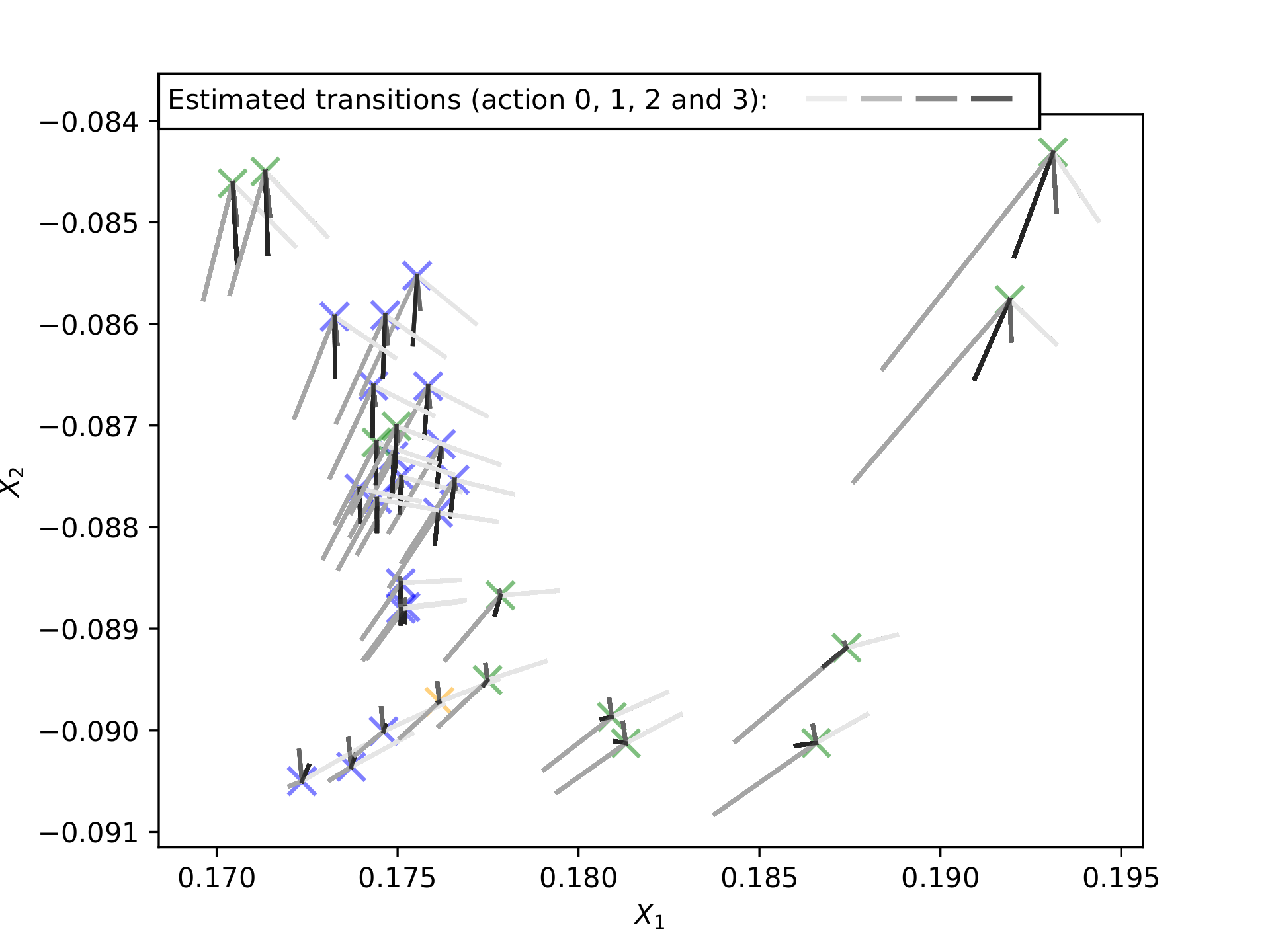}
  \caption{Ablation of the representation loss $\mathcal L_{d}$ and replacement by an auto-encoder loss.}
  \label{fig:lab_dis_abla6}
    \end{subfigure}
    \caption{Study of the importance of the representation loss $\mathcal L_{d}$ that prevents all representations to collapse.}
\end{figure}

In order to illustrate the importance of the representation loss $\mathcal L_{d}$, we first perform the same experiment as in Figure \ref{fig:lab_dis} but we do not update the weights of the encoder to optimize the loss $\mathcal L_{d}$. As can be seen in Figure~\ref{fig:lab_dis_abla5}. Without the entropy maximisation at the output of the encoder, all representations tend to collapse to a constant point in the low-dimensional abstract space, thus losing the usefulness of the approach.

One can wonder if it's possible to use an auto-encoder loss instead of the representation loss $\mathcal L_{d}$. It can be seen in Figure \ref{fig:lab_dis_abla6} that, in this case, the auto-encoder loss is not sufficient to ensure a sufficient diversity in the the low-dimensional abstract space and all abstract representations tend still to collapse to a constant point (even though less dramatically than in Figure \ref{fig:lab_dis_abla5}). This happens because a sufficiently rich decoder can reconstruct the input even though the abstract representations have almost collapsed to a constant point. Optimizing both an auto-encoder loss and a transition loss $\mathcal L_\tau(\theta_e, \theta_\tau)$ does therefore not lead to a robust solution to ensure the diversity in the low-dimensional abstract state space. 
For that experiment, the decoder uses the same architecture than the encoder (in a reverse order). The loss of the auto-encoder is the $L_2$ reconstruction loss and the same learning rate is used.

\section{Details on the catcher environment}
\label{app:catcher}
The catcher environment considered in this paper is made up of balls (blocks of dark pixels) that periodically appear at the top of the frames (at random horizontal positions) and fall towards the bottom of the frames where a paddle has the possibility to catch them.
The agent has two possible actions: moving the paddle in the directions left or right (by three pixels). At each step the ball falls (also by three pixels) and when the ball reaches the bottom of the screen, the episode ends and the agent receives a positive reward of +1 for each ball caught, while it receives a negative reward  -1 if the ball is not caught. Only two possible starting positions for the ball are considered (either on the far left or the far right of the screen).

The following learning rates are used:
$\alpha=5 \times 10^{-4}$ (decreasing by 10\% every 2000 training steps);
$\beta=0.2$;
$\alpha_{interpr}=\alpha$ (when used).

The experiments on that domain are obtained in an online setting context. A new sample is obtained via a random policy at every step and kept in a replay memory.

\section{Details on the distribution of labyrinths}
\label{app:distrib}
The state representation is a 2 dimensional array of $48 \times 48$ elements $\in [-1,1]$ (one channel with grey-levels are provided as input).
For the distribution of labyrinth, the underlying dynamics happen in a $8 \times 8$ grid. The grid has walls on all its contour and the agent starts in the top left corner. The walls and the rewards are added randomly:
\begin{itemize}
\item 16 walls are placed randomly on the grid (in addition to the contour), and
\item 3 keys are also added randomly on free positions.
\end{itemize}
A rejection step is performed for all labyrinths for the cases where the agent has not the possibility of reaching all three keys.

The discount factor is $G(s,a,s')=1$ for all transitions from state $s$ to state $s'$ with action $a$, except when all the keys have been gathered by the agent, in which case $G(s,a,s')=0$. The biased discount factor used for training is $\Gamma(s,a,s')=\text{argmin}(G(s,a,s'),0.9)$.

When gathering trajectories with the random policy, a new labyrinth is sampled from the distribution only once all rewards have been gathered by the agent. In the test phase, the episode is terminated either once all rewards have been gathered or when the number of 50 steps has been reached (thus the lowest score for an episode is -5 in the case where no reward has been gathered).

The policy follows the optimal action given by
$\underset{a \in \mathcal A}{\operatorname{argmax}} \ Q_{plan}^D(\hat x_t, a)$ 90\% of the time and a random action 10\% of the time. When planning to estimate $Q_{plan}^D(\hat x_t, a)$, as described in Section \ref{sec:planning}, only b-best options are considered. For the first planning step, all four best actions are kept and in the following only the two best actions are kept at each expansion step. 

The following hyper-parameters are used:
$\alpha=5 \times 10^{-4}, \beta=1$.

\subsection{Sensibility study to hyper-parameters}

\begin{figure}[ht!]
    \centering
    \begin{subfigure}[t]{0.22\textwidth}
  \centering
  \includegraphics[width=.99\linewidth]{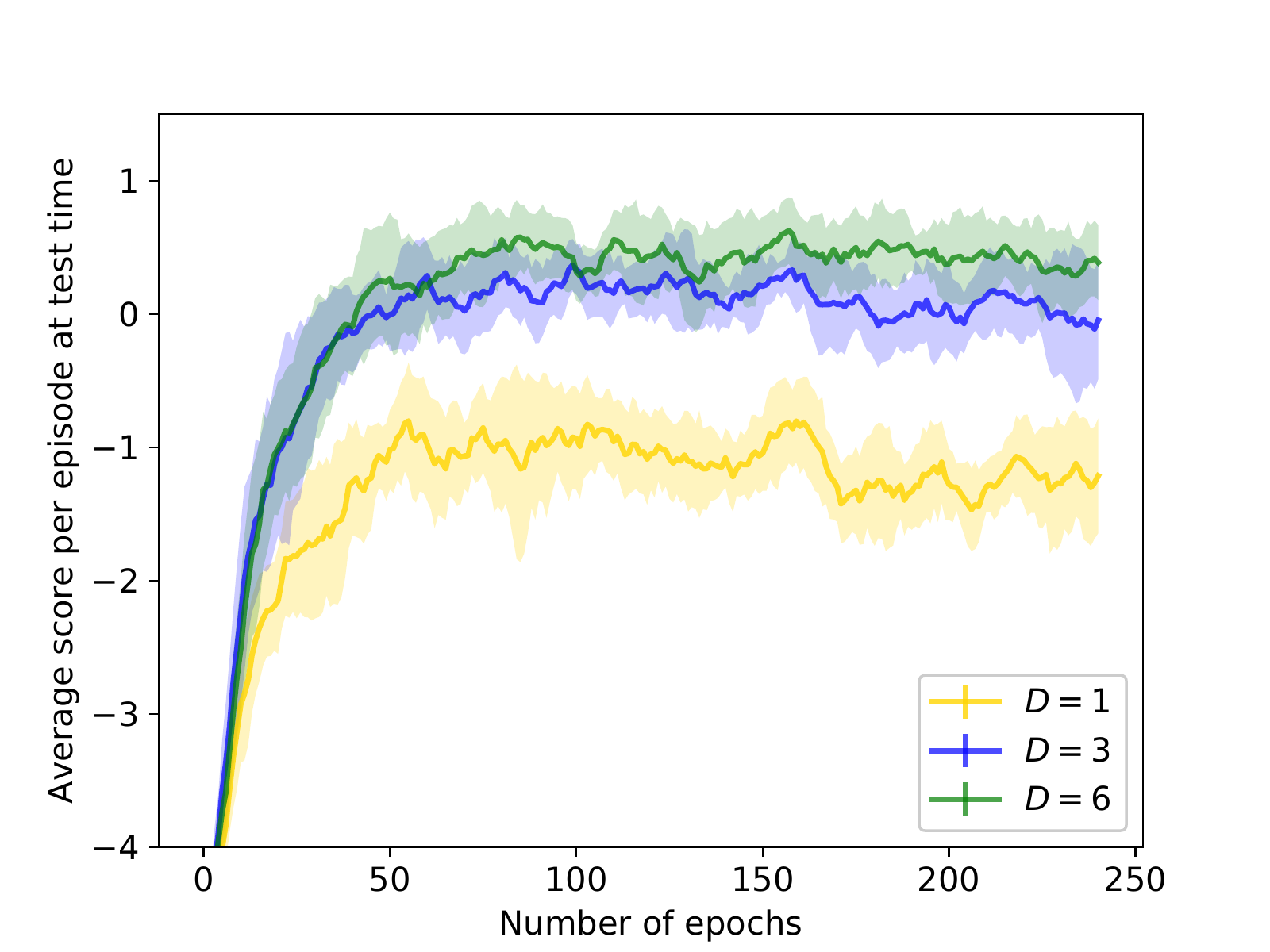}
\caption{The setting is the same as in Figure \ref{fig:meta}, except that training is done with $10^5$ transitions obtained off-line by a random policy.}
\label{fig:meta_abla1}
    \end{subfigure}%
    \ \ \
    \begin{subfigure}[t]{0.22\textwidth}
  \centering
  \includegraphics[width=.99\linewidth]{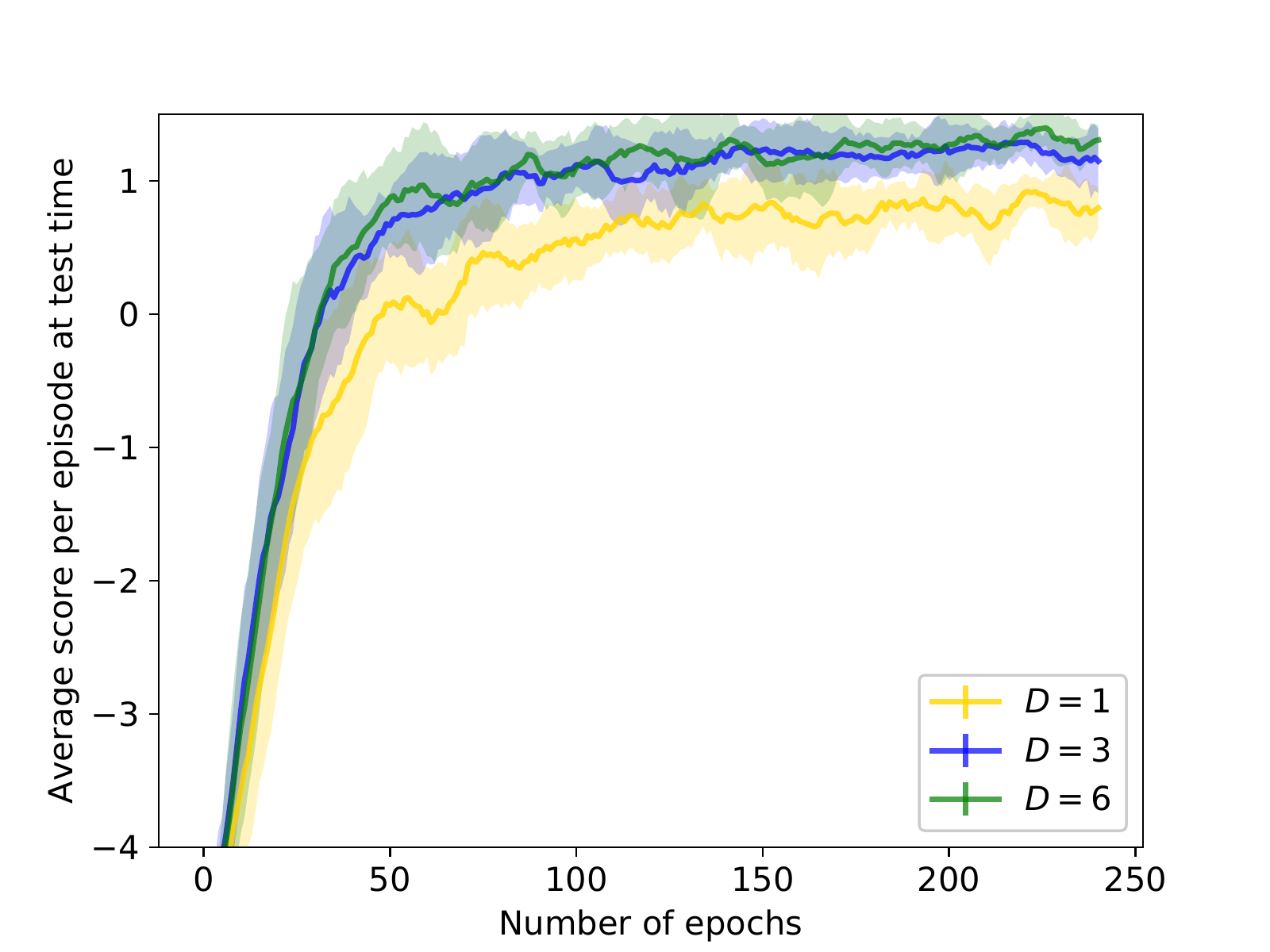}
\caption{The setting is the same as in Figure \ref{fig:meta}, except that training is done with $5 \times 10^5$ transitions obtained off-line by a random policy.}
\label{fig:meta_abla2}
    \end{subfigure}
    \caption{Meta-learning score on a distribution of labyrinths.}
\end{figure}

On the one hand, when a high amount of data is available, planning is not crucial (see Figure \ref{fig:meta_abla2}). On the other hand, when the amount of data available for the task decreases, the combination of model-based and model-free becomes more important (see Figure \ref{fig:meta_abla1}).

\end{document}